\documentclass[10pt, a4paper]{article}  

\usepackage[english]{babel}   
\usepackage[utf8]{inputenc}   
\usepackage[T1]{fontenc}      
\usepackage{lmodern}          

\usepackage[a4paper,margin=1in]{geometry}
\usepackage{authblk}          
\usepackage{setspace}

\usepackage{amssymb, amsmath, mathtools}  
\usepackage{amsthm}           
\usepackage[mathscr]{euscript} %

\usepackage[ruled,linesnumbered]{algorithm2e}  

\usepackage{caption}          
\usepackage{subcaption}       
\usepackage{xspace}
\usepackage{xcolor}
\usepackage{graphicx}         
\usepackage{colortbl}

\usepackage{tabularx}
\usepackage{booktabs}
\usepackage{siunitx}

\usepackage[title]{appendix}

\definecolor{moplorange}{rgb}{0.96, 0.82, 0.66}
\definecolor{moporange}{rgb}{0.92, 0.6, 0.28}
\definecolor{moplblue}{rgb}{0.10, 0.48, 0.86}
\definecolor{mopblue}{rgb}{0.07, 0.35, 0.63}
\definecolor{mopdblue}{rgb}{0.05, 0.25, 0.45}
\definecolor{mygreen}{rgb}{0.058,.596,0.047}  
\colorlet{mopsorange}{moporange!60!red}
\colorlet{mopsblue}{moplblue!60!blue}
\colorlet{mopsgreen}{mygreen!60!green}

\usepackage{hyperref}         
\usepackage[dvipsnames]{xcolor}    
\hypersetup{
  colorlinks=true,            
  linkcolor=blue,             
  urlcolor=purple,            
  citecolor=teal              
}

\usepackage{csquotes}               
\usepackage[round]{natbib}
\usepackage{enumitem}               
\setlist[itemize]{left=0pt}         
\setlist[enumerate]{left=0pt}       

\newcommand{\idmat}{I}                      

\newcommand{\fielduns}{v_t^{1}(x)}
\newcommand{\fieldunscont}{v_t^{1,\epsilon}(x)}
\newcommand{\fieldsquared}{v_t^{2}(x)}
\newcommand{\fieldsquaredcont}{v_t^{2,\epsilon}(x)}

\newcommand{\posteriorQuot}{\Gamma_t(x)}

\DeclareMathOperator{\diver}{div}

\DeclareMathOperator*{\argmin}{arg\,min}

\newtheorem{theorem}{Theorem}[section]          
\newtheorem*{theorem*}{Theorem}  
\newtheorem{lemma}[theorem]{Lemma}              
\newtheorem*{lemma*}{Lemma}
\newtheorem{corollary}[theorem]{Corollary}      
\newtheorem*{corollary*}{Corollary} 

\title{\textbf{Euclidean Distance-based Flow Matching}}

\author{Paul-Henrik Heilmann\thanks{Email: \href{mailto:heilmann@math.uni-sb.de}{heilmann@math.uni-sb.de}} }
\author{Peter Ochs\thanks{Email: \href{mailto:ochs@math.uni-sb.de}{ochs@math.uni-sb.de}}}

\affil{Saarland University, Saarbrücken, Germany}

\date{}  

\begin{document}

\maketitle  


\begin{abstract}
  Flow Matching has recently become popular as a technique for generative modeling with applications in many domains. In the standard Flow Matching formulation, we interpret the vector field as the solution of a least-squares regression problem, which is known to be sensitive to contamination. Therefore we leverage the viewpoint of robust statistics and replace the squared Euclidean distance with the Euclidean distance as a regression objective. This seemingly minor change strongly affects the properties of Flow Matching. We provide a probabilistic interpretation of the new regression target as a conditional geometric median and use this characterization to study distributional robustness of the Euclidean Distance-based Flow Matching. We analyze the mathematical properties of this changed objective and the effect of a contaminated target distribution on the involved conditional velocity distributions. We discuss the challenge of directly characterizing the distribution that our method induces. In specific contamination settings we show theoretically that our method yields an improvement for severe contamination. Finally, we present numerical experiments conducted on the CIFAR-10 image dataset where a proportion of the samples has been contaminated. Our Euclidean distance-based Flow Matching shows a clear advantage in scenarios of severe contamination of the target distribution. 
\end{abstract}

\section{Introduction}
Flow Matching (FM) \citep{lipman2023flowmatchinggenerativemodeling,lipman2024flowmatchingguidecode,albergo2023buildingnormalizingflowsstochastic,liu2022flowstraightfastlearning} is a framework for training a flow model \citep{NEURIPS2018_69386f6b,grathwohl2018ffjord} to generate samples from a target distribution that is available only through data. Starting from a known source distribution from which we can sample, FM learns a time-dependent vector field that drives the generation process. New samples are generated by drawing from the source distribution and applying the flow induced by this learned time-dependent vector field.
It is well known that under suitable regularity assumptions, the (population) optimal vector field for the standard formulation generates the prescribed target distribution exactly.

Real-life datasets often do not only contain samples from the assumed true distribution, but they contain contaminated samples as well. A classical example is the Belgium telephone calls dataset  \citep{rob_outlier} that includes several wrongly recorded data points. Therefore we model the observed distribution as a mixture model of the clean and a contaminating distribution, rather than only the target distribution. Consequently, robustness is the desirable property that the learned distribution stays close to the clean distribution, even in the presence of contamination without manually filtering the samples. 

Flow Matching is a powerful technique to produce samples from the target, but suffers from replication of the exact target including the contamination. We present a remedy for this by a simple modification of the Flow Matching objective. 
Our approach replaces the squared Euclidean distance loss in the standard Flow Matching objective with the Euclidean distance loss. The latter penalizes large regression residuals linearly rather than quadratically, motivating its use to reduce sensitivity to contamination. At the population level, the modified field is a geometric median of the conditional velocity distribution, whereas the standard squared-loss field is its mean. We demonstrate that this change also improves distributional robustness, yielding a generated distribution closer to the clean target under contamination. Clearly the necessary trade-off that we accept is that we lose the general guarantee that the population-optimal field reproduces the prescribed target distribution, even when the data are uncontaminated.

To be more precise, we characterize the population-optimal Euclidean distance Flow Matching vector field as a conditional geometric median. On clean data, we derive bounds comparing this field with the standard squared-loss field.  We then analyze how contamination affects the regression fields and prove that, in specific contamination settings, our adaptation yields an improvement in the Wasserstein distance. Finally, we evaluate both objectives on contaminated CIFAR-10 data, comparing their generation quality across different contamination models and proportions.

\subsection{Related Work}
Several works focus on improving Flow Matching by modifying the coupling between source and target samples, for instance \citet{pooladian2023multisampleflowmatchingstraightening} and \citet{tong2024improving}. Repeated rectification also straightens trajectories \citep{liu2022flowstraightfastlearning}. These approaches modify the coupling or transport path while retaining the standard squared regression objective.
\citet{l1sampleflow} consider a related alternative objective in visuomotor learning. In their main method, they reformulate velocity prediction as terminal-sample prediction and train the resulting model using a componentwise absolute-deviation loss together with a two-step sampling procedure, although they also test their loss with conventional flow integration. They empirically demonstrate faster convergence on visuomotor-learning benchmarks. We use Euclidean-distance based loss, which is invariant under orthogonal changes of coordinates and yields a conditional geometric median rather than coordinatewise conditional medians. Moreover, our focus is distributional robustness.

Robustness was recently studied in the closely related setting of diffusion models and score matching. \citet{khrapov2024improvingdiffusionmodelssdatacorruption} replace the standard squared denoising objective with a scheduled pseudo-Huber loss with a time-varying threshold. They empirically demonstrate improved resistance to contaminated training data in both image and speech generation tasks. Similarly \citet{li2026robustdiffusionmodelsdivergenceinduced} study replacing the mean squared error denoising loss with transformations induced by f-divergences. Their framework includes bounded-influence choices that downweight samples with large residuals and empirically improve generation under contamination. In the distinct setting of parametric score matching, \citet{pmlr-v258-schwank25a} construct a robust estimator through geometric median of means aggregation and establish theoretical guarantees with particular attention to exponential-family models. Our approach differs in that the  geometric median is not introduced as an external aggregation procedure but instead arises directly as the population-optimal regression target of our modified objective. 

\section{Technical Background}
Let us briefly recall technical foundations of Flow Matching.
Denote the source distribution as $P_0$ and the target distribution as $P_1$. Throughout we assume the source to be Gaussian $P_0 = \mathcal{N}(0,\idmat_n)$. We follow the interpolation path $(P_t)_{t \in [0,1]}$ which is the linear path interpolating independent random vectors \citep{peyré2025optimaldiffusiontransportsmachine}:
\begin{equation*}
    P_t=\textup{Law}(X_t), \quad \textup{where} \quad(X_0,X_1) \sim P_0 \otimes P_1 \, ,  X_t = (1-t)X_0 +tX_1.
\end{equation*}
Throughout we assume that all endpoint distributions have finite second moments.
The goal is to find a time dependent vector field $v:  [0,1] \times \mathbb{R}^n \to \mathbb{R}^n, \, (t,x) \mapsto v_t(x) $ such that the pair $(P_t , v_t)$ satisfies the continuity equation (technically understood in the weak sense)
\begin{equation*}
    \partial_t P_t+ \diver(v_t P_t) = 0.
\end{equation*}
The flow $\phi : [0,1] \times \mathbb{R}^n \to \mathbb{R}^n$ induced by $v_t(x)$ is defined as the solution to the following ordinary differential equation (ODE)
\begin{equation*}
    \frac{\textup{d}}{\textup{d}t} \phi_t(x) = v_t(\phi_t(x)), \quad \textup{with } \phi_0(x) = x.
\end{equation*}

If $v_t(x)$ is sufficiently regular it follows from \citet{bookVillaniTransport} that its induced flow generates the previously defined probability path $P_t$ in the sense that $(\phi_t)_\# P_0 = P_t$. In particular, it follows that $\phi_1(X_0) \sim P_1$. Consequently such a vector field enables generating samples from the target distribution, by sampling from the source and integrating the ODE which is done numerically. In standard Flow Matching the population objective is 
\begin{equation}\label{minProb}
    \min_{(v_t)_{t\in [0,1]}} \int_0^1 \int_{\mathbb{R}^n \times \mathbb{R}^n} \lVert v_t((1-t)x_0 + tx_1) - (x_1-x_0) \rVert^2 \textup{d} P_0(x_0) \textup{d}P_1(x_1) \textup{d}t.
\end{equation}
Throughout the paper, we denote by $\fieldsquared$ a measurable population minimizer of \eqref{minProb}. By the optimal regression property under a squared objective, we obtain that the population-level minimizer can be written as a conditional expectation \citep{peyré2025optimaldiffusiontransportsmachine,albergo2025stochasticinterpolantsunifyingframework}:
\begin{equation}\label{exp:cond:inter}
    \fieldsquared = \mathbb{E} \left[ X_1 - X_0 \vert X_t=x \right].
\end{equation} Then $\fieldsquared$ and $P_t$ solve the continuity equation in the weak sense. 
In practice we work with \eqref{minProb} which is tractable via empirical risk minimization. Using a simple source distribution we can generate samples from $P_0$ and the target distribution is available through data samples.
\subsection{Euclidean distance regression objective}
We aim to study the effects of replacing the squared Euclidean norm in the regression problem $ \lVert \cdot \rVert^2$ with the norm $\lVert \cdot \rVert$. We investigate whether the reduced sensitivity of this regression target to contamination can yield a terminal distribution closer to the clean target when the observed target distribution is contaminated. \citet{lipman2024flowmatchingguidecode} show that the optimal vector field after replacing the loss $\lVert \cdot \rVert^2 $ in \eqref{minProb} by a Bregman divergence, applied with the target velocity as the first argument and the prediction as the second, preserves the probability path. Our proposed loss is not covered by this and we have no guarantee that our method generates the probability path $P_t$. In particular, we cannot guarantee that the learned target coincides with the observed target. We highlight that this is an important property of our method, since we do not want to replicate contaminated distributions. 

Removing the square in \eqref{minProb} our regression objective becomes 
\begin{equation}\label{unsquared_reg}
    \min_{(v_t)_{t \in [0,1]}} \int_0^1 \int_{\mathbb{R}^n \times \mathbb{R}^n} \lVert v_t((1-t)x_0 + tx_1) - (x_1-x_0) \rVert \textup{d} P_0(x_0) \textup{d}P_1(x_1) \textup{d}t.
 \end{equation}
We denote by \(\fielduns\) a measurable population minimizer of
\eqref{unsquared_reg}.
The integral can be seen as a mixed norm and consequently we will also call the method $L_{2,1}$ Flow Matching. 
\subsection{Discussion of mathematical details} All of our proofs can be found in the supplementary material \ref{appen:proofs}.
Since conditional distributions are unique only $P_t$-almost everywhere, characterizations of vector fields as quantities of conditional distributions, as in \eqref{exp:cond:inter}, only hold for Lebesgue-almost every $t$ and $P_t$-almost every $x$. The same holds for all pointwise identities and inequalities involving these vector fields. When the conditional geometric median, which is introduced below, is not unique, $\fielduns$ denotes a chosen measurable population minimizer.
Note that even when the conditional-mean field admits a continuous representative, the same does not need to hold for the geometric-median field. Although the almost-everywhere characterizations are sufficient for the population regression problems, they do not by themselves identify regular representatives whose associated ODEs are well posed. The existence and uniqueness of induced flows are separate questions that require choosing suitable representatives and imposing additional regularity assumptions. Whenever we analyze an induced distribution, we therefore either construct a particular measurable representative explicitly or state the required existence and regularity assumptions. We make no general claims concerning ODE well-posedness for the \(L_{2,1}\) population minimizer.
\section{Geometric Median Characterization}
In this section we first introduce the geometric median and use it to characterize the population minimizer for the $L_{2,1}$-Flow Matching objective. The geometric median properties used in this section are classical and we include their proofs for convenience. We then apply these properties of the geometric median to help us to compare the resulting vector field with the standard Flow Matching vector field. In one dimension, the median is a measure of location that is more robust to contamination than the mean. The generalization to higher dimensions is not straightforward and multiple generalizations have been suggested \citep{1f6117c1-50e3-3658-911b-2cdfe87c4f25}. One possible way to do so is the geometric median \citep{10.1093/biomet/35.3-4.414,Minsker_2015}.
For a probability measure $\mu$ on $\mathbb{R}^n$ with finite first moment define its geometric median as
\begin{equation*}
     \textup{Gmed}(\mu) = \argmin_{y \in \mathbb{R}^n} \int_{\mathbb{R}^n}\lVert y-x \rVert  \mu (\textup{d}x).
\end{equation*}
Its existence is established in \citet{Minsker_2015}. If the distribution is not concentrated on an affine line, this object is unique \citep{10.1214/aos/1176350511}. We will give a probabilistic interpretation of the solution of \eqref{unsquared_reg}, similar to the mean characterization of squared Flow Matching. For a random variable $X$ we write 
$$\operatorname{Gmed}(X) \coloneqq \operatorname{Gmed}(\operatorname{Law}(X)) $$
and similarly for a conditional distribution, we write
  $$  \operatorname{Gmed}(X\mid Z=z)
    \coloneqq
    \operatorname{Gmed}(\operatorname{Law}(X\mid Z=z)).$$

\subsection{Comparison with standard Flow Matching}\label{subsec:vectorfields}
Now we want to connect the geometric median to our proposed Flow Matching method and analyze the difference in the resulting vector fields.
\begin{lemma}\label{lemma:field_gmed}
 It holds that
    \begin{equation*}
        \fielduns \in \textup{Gmed}(X_1-X_0 \vert X_t = x),
    \end{equation*}
    for almost every $t$ and $P_t$-almost every $x$. Moreover, if this conditional distribution is not concentrated on an affine line for almost every $t$ and $P_t$-almost every $x$, then the minimizer is unique up to almost everywhere equality. 
\end{lemma}
While this result yields a probabilistic interpretation of the $L_{2,1}$-regression objective, we are unable to give a characterization of the general induced probability path. The proof for standard Flow Matching relies on the tower property of conditional expectations, which to our knowledge does not have a counterpart for the geometric median. Consequently, the standard Flow Matching argument does not show that the $L_{2,1}$-population field generates the prescribed interpolation, and our analysis does not provide a general explicit description.

In the following we analyze the difference between the two vector fields for special scenarios. 
We first derive that for symmetric distributions, the two vector fields coincide.
\begin{lemma}\label{lem:symmetry}
Assume that, for almost every $t$ and $P_t$-almost every $x$,
the conditional distribution of $X_1-X_0$ given $X_t=x$ is not concentrated on an affine line and is centrally symmetric around its mean, i.e., for any random variable having this distribution, say Y, we have that  $ Y-\mathbb{E}[Y]$ matches, in distribution, $-(Y-\mathbb{E}[Y])$. Then we have that
        \begin{equation*}
            \fielduns = \fieldsquared
        \end{equation*}
        for almost every $t$ and $P_t$-almost every $x$.
\end{lemma}
Generally, the target distribution being symmetric around its mean does not imply that the same holds for the conditional velocity distribution of $X_1 -X_0 $ given $X_t = x$, even though the Gaussian source is symmetric. As a sufficient condition, if the target distribution is Gaussian, then we can show that the conditional velocity distributions are also Gaussian and thus we obtain the following: 

\begin{corollary}\label{corr:gauss:equality}
    If the source distribution is a standard Gaussian distribution $X_0 \sim \mathcal{N}(0, \idmat)$ and the target distribution is nondegenerate Gaussian $X_1 \sim \mathcal{N}(\mu, \Sigma)$ where $X_0$ and $X_1$ are independent, then \begin{equation*}
            \fielduns = \fieldsquared
        \end{equation*}
        for almost every $t$ and $P_t$-almost every $x$.
\end{corollary}
The setting of Corollary \ref{corr:gauss:equality} seems to be quite special however, it is known to be a good proxy in high-dimensional statistics. 
More generally, the integrated squared difference between the two vector fields is bounded by the optimal squared Flow Matching regression loss.
\begin{theorem}\label{thm:cond_vel}
In our $L_{2,1}$ Flow Matching setup we have that:
    \begin{equation*}
        \int_0^1 \int_{\mathbb{R}^n} \lVert \fielduns - \fieldsquared \rVert^2 P_t(\textup{d}x) \textup{d}t \leq \int_0^1 \mathbb{E}\lVert (X_1-X_0) - v_t^{2} (X_t) \rVert^2 \textup{d}t
    \end{equation*}
    where the right-hand side is the optimal squared FM regression loss. 
\end{theorem}

Intuitively this shows that the better the velocity can be predicted from interpolation points, the closer the two fields are to each other. This derived velocity field bound is generally not sharp. For instance, for Gaussian targets we have established that both target vector fields are equal, while our bounds do not reflect this. Under additional regularity assumptions, for instance if $\fielduns$ admits a globally Lipschitz representative the above vector field bound could be translated into a Wasserstein bound between the induced distributions. We do not pursue such flow level estimates here.

\section{Robustness analysis}\label{sec:rob_ana}
In this section, we study how contamination of the target distribution affects the population vector fields of both the squared and the $L_{2,1}$ method.
The robustness of the geometric median has been assessed in various works \citep{1f6117c1-50e3-3658-911b-2cdfe87c4f25}. It also appears in more recent works, as an aggregate estimator \citep{Minsker_2015,doi:10.1137/23M1592420}.
However, in our Flow Matching setting the geometric median is only applied on a conditional distribution at every time and space position to yield a vector field, whose integration in turn yields the generative model. Therefore the robustness properties are highly non-trivial.
We separate our analysis into two levels. First, we study how target contamination affects the conditional velocity distributions and the corresponding population vector fields. Second, we investigate whether this robustness can translate into a generated distribution that is closer to the clean target.

We consider a Huber contamination model \citep{10.1214/aoms/1177703732}, i.e. we assume that the true target distribution is $ P $  and a proportion of $\epsilon$ observations are instead replaced with a contaminating distribution $Q$:
\begin{equation*}
    P^{\textup{cont}} = (1-\epsilon)P + \epsilon Q, \quad 0 \leq \epsilon < 1/2.
\end{equation*}
A crucial property is given by the geometric median's so-called high finite-sample breakdown point. That is the smallest proportion of observations that must be replaced by arbitrary values to make an estimator arbitrarily large or otherwise uninformative. Its finite-sample replacement breakdown point is $ B_{\textup{GM}} =\lfloor  (N+1)/2\rfloor /N$ \citep{article}. This means that if a proportion smaller than $B_{\textup{GM}}$ of the $N$ observations is replaced, then the displacement of the geometric median is bounded, regardless of the strength of the contamination. In contrast, replacing even a single observation by a point arbitrarily far from the rest of the data can cause an arbitrarily large displacement of the sample mean. This property motivates our use of the geometric median as a velocity estimator, as it remains stable under a substantial proportion of arbitrarily corrupted observations. However, this property does not mean that the geometric median yields better results in all contamination settings. In particular, the mean of a contaminated distribution may be closer to the clean mean than the geometric median.

\subsection{Robustness properties of the geometric median field}
We first study how contamination of the target distribution changes the conditional velocity distributions and the corresponding standard and $L_{2,1}$-vector fields. Since population minimizers are defined only up to almost-everywhere equivalence, we fix measurable representatives \(v_t^2,v_t^1\) for the clean target and \(v_t^{2,\epsilon},v_t^{1,\epsilon}\) for the contaminated target. Let
\begin{equation*}
    B \sim \operatorname{Bernoulli}(\epsilon) ,  X_P \sim P, X_Q \sim Q,X_0 \sim P_0 
\end{equation*}
be mutually independent.  Let 
\begin{equation*}
    X_1^{\textup{cont}} = (1-B)X_P + B X_Q, \quad X_t^{\textup{cont}} = (1-t)X_0 + tX_1^{\textup{cont}},
\end{equation*} and
\begin{equation*}
  X_t^P = (1-t)X_0 + tX_P\quad \, ,  X_t^Q = (1-t)X_0 + tX_Q  .
\end{equation*}
For $t$, let $\gamma_t$ be a measurable version of the posterior contamination probability
\begin{equation*}
    \gamma_t(x) = \mathbb{P} \left[ B=1 \vert X^{\textup{cont}}_t = x \right].
\end{equation*}
Intuitively, this quantifies the likelihood that after we have sampled $t$ and the corresponding interpolation point, the sampled endpoint was caused by the contaminating distribution. Using the law of total probability we obtain
\begin{align}
&\operatorname{Law}\left(
    X_1^{\mathrm{cont}}-X_0
    \,\middle|\,
    X_t^{\mathrm{cont}}=x
\right)  \nonumber \\
&\quad=
(1-\gamma_t(x))
\operatorname{Law}\left(
    X_P-X_0
    \,\middle|\,
    X_t^P=x
\right)
+
\gamma_t(x)
\operatorname{Law}(
    X_Q-X_0
    \,|\,
    X_t^Q=x
). \label{eq:prob_decom}
\end{align}
This is again a mixture distribution, but its weight is the posterior probability $\gamma_t(x)$ rather than $\epsilon$. Moreover, $\gamma_t(x)$ depends on the geometry and magnitude of the contamination. Consequently, moving the contamination farther from the clean data does not uniformly improve the relative robustness of the geometric median. When \(\gamma_t(x)\geq 1/2\), the clean conditional component no longer has a strict majority, so the robustness guarantee below does not apply. Using the linearity of the mean, we can characterize the influence of contamination on the squared vector field
\begin{equation*}
    \lVert \fieldsquaredcont-\fieldsquared\rVert
    =
    \gamma_t(x)
    \lVert
        \mathbb{E}[
            X_Q-X_0
            \,\vert\,
            X_t^Q = x
        ]
        -
        \mathbb{E}[
            X_P-X_0
            \,\vert\,
            X_t^P=x
        ]
    \rVert.
\end{equation*}

We notice that in the norm, only the clean conditional mean is independent of the contamination,  whereas both $\gamma_t(x)$ and the contaminated conditional mean depend on the contamination. The difference between the conditional means may become arbitrarily large as the contamination is moved farther from the clean data. To fully understand this, one would need to control the posterior contamination probability $\gamma_t(x)$.
On the other hand, for $\gamma_t(x) < 1/2 $ we have the following bound, adapted from a general result for the geometric median by \citet{Minsker_2015}.
\begin{lemma}\label{unsq_cont_bound}
For almost every $t$ and $P_t^\textup{cont} $-almost every $x$ satisfying  $\gamma_t(x) < 1/2$, there exists $C_t(x) \ge 0$, depending only on the clean model and not on the contamination, such that
\begin{equation*}
     \lVert \fieldunscont - \fielduns \rVert \leq \frac{2(1-\gamma_t(x))}{1-2\gamma_t(x)} C_t(x).
\end{equation*}    
\end{lemma}

For $ \gamma_t(x) < 1/2$ the factor $\frac{2(1-\gamma_t(x))}{1-2\gamma_t(x)}$ is positive and finite, although it diverges as $\gamma_t(x) \to 1/2$. Generally, those bounds do not directly compare to each other. The squared vector field identity depends on the outlier magnitude and the posterior contamination probability. The $L_{2,1}$-vector field bound depends on a constant that is independent of the contamination and the posterior contamination probability. This suggests that the $L_{2,1}$-vector field can be less distorted by remote contamination whenever $\gamma_t(x)<1/2$, but it does not imply that it is uniformly closer to the clean field.

Since $v_t^{2}$ solves the continuity equation with the clean probability path, we are interested in how much the contaminated $L_{2,1}$-vector field deviates from it. Using the triangle inequality gives
\begin{equation*}
    \lVert \fieldsquared - \fieldunscont \rVert 
    \leq \lVert \fieldsquared  -\fielduns \rVert + \lVert  \fielduns -  \fieldunscont \rVert.
\end{equation*}
This decomposition separates the discrepancy of the mean and the median from the additional distortion caused by contamination. In Section \ref{subsec:vectorfields} we discussed bounds for the first term, while in this section we have established under which conditions we can control the second term. Since this is usually not enough to provide distributional guarantees, we will now present two special settings where we can control the distributional error. 
\subsection{Wasserstein bounds for special cases}
To quantify the discrepancy between generated and clean distributions, we use the $1$-Wasserstein distance. For probability measures $\mu$ and $\nu$ on $\mathbb{R}^n$ with finite first moments, it is defined as
\begin{equation*}
    W_1(\mu,\nu)
    \coloneqq
    \inf_{\pi\in\Pi(\mu,\nu)}
    \int_{\mathbb{R}^n\times\mathbb{R}^n}
    \lVert x-y\rVert\,\pi(\mathrm{d}x,\mathrm{d}y),
\end{equation*}
where $\Pi(\mu,\nu)$ denotes the set of couplings of $\mu$ and $\nu$. Throughout this subsection, since we use linear interpolation with Gaussian source, for an endpoint distribution $\mu$ we fix the posterior 
\begin{equation*}
    \pi_{t,x}^\mu (\textup{d}y) \propto \exp \left( \frac{-\lVert x-ty \rVert^2}{2(1-t)^2} \right) \mu(\textup{d}y), \quad (t,x) \in [0,1) \times \mathbb{R}^n,
\end{equation*}
where the right-hand side is normalized to unit mass. This choice allows us to impose the geometric median condition for every $t,x$. Then given a Borel measurable $m^\mu$ satisfying $m_t^\mu (x) \in \operatorname{Gmed}(\pi_{t,x}^\mu)$ for every $(t,x) \in [0,1) \times \mathbb{R}^n$ the corresponding pointwise $L_{2,1}$ velocity field is 
\begin{equation}\label{concrete:def:field}
    v_t^{1,\mu}(x) = \frac{m_t^{\mu}(x) -x}{1-t}.
\end{equation}
Throughout this subsection, solutions of the associated ODE are understood in the sense that their paths are locally absolutely continuous and satisfy the ODE almost everywhere on \([0,1)\).
We first consider a point-mass clean target with point-mass contamination and show that our proposed method yields a smaller Wasserstein distance to the clean target in this setting.
\begin{theorem}\label{thm:dirac}
Let $\delta_0 $ be a Dirac distribution in $\mathbb{R}^n$ centered at $0$. Let $0 < \epsilon<1/2$. For $b \in \mathbb{R}^n \setminus \{0\}$, consider the contaminated distribution
\begin{equation*}
    P^{\textup{cont}} = (1-\epsilon)\delta_0 + \epsilon \delta_b.
\end{equation*}
For every pointwise $L_{2,1}$-velocity field $v^{1,P^{\textup{cont}}}$ coming from a Borel geometric-median selection as defined in \eqref{concrete:def:field}, the associated ODE admits a solution process $(Z_t)_{t\in[0,1)} $ with $Z_0 \sim \mathcal{N}(0,\idmat_n)$. Every such solution converges almost surely as $t \nearrow 1 $ and its terminal law $\rho_1 = \textup{Law}(\lim_{t \nearrow1} Z_t)$ satisfies
    \begin{equation*}
        W_1(\rho_1,\delta_0)  < W_1(P^{\textup{cont}},\delta_0).
    \end{equation*}
\end{theorem}
For a point-mass clean target, the preceding theorem establishes improvement for every nonzero displacement of the contaminating point. We now present our main theoretical result, which establishes a corresponding improvement for bounded clean target distributions when the contaminating point is sufficiently far away.

\begin{theorem}\label{thm:bd}
Let \(P\) be a distribution supported in the closed ball \( \overline{B}_1(0)\). Let \(0<\epsilon<1/2\), and fix a unit vector \(u\in\mathbb R^n\). For \(R>1\), set \(b_R=Ru\) and
$$
P_R^{\mathrm{cont}}
=
(1-\epsilon)P+\epsilon\delta_{b_R}.
$$
For each \(R>1\), let \(v^{1,R}\) be any pointwise \(L_{2,1}\)-velocity field associated with \(P_R^{\mathrm{cont}}\) as defined in \eqref{concrete:def:field}. Assume that its associated ODE admits a solution process with initial distribution \(\mathcal N(0,I_n)\). Let \((Z_t^R)_{t\in[0,1)}\) be any such solution process, and write $\rho_t^R=\operatorname{Law}(Z_t^R).$ Then, for all sufficiently large \(R\),
$$
\limsup_{t\nearrow1}
W_1(\rho_t^R,P)
<
W_1(P_R^{\mathrm{cont}},P).
$$

\end{theorem}
Standard FM aims to reproduce the observed target distribution, which in our setting includes contamination.
Our theorems establish that the robustness of the geometric median can translate into a smaller distributional error, in comparison to the contaminated target. For a point-mass target, improvement holds for every nonzero contaminating point. Our main result establishes improvement for bounded target distributions if the contaminating point is sufficiently far away. In this more general setting, the limit superior means that the improvement holds for all times  sufficiently close to the final time. This formulation does not require convergence to the final distribution.

\section{Experiments}\label{sec:experiments}
We compare classical Flow Matching with our proposed formulation under clean and contaminated distributions. Specifically, we investigate how our $L_{2,1}$-Flow Matching retains the performance of classical Flow Matching on clean data, whether it is more robust under contamination, and whether it depends on the proportion and type of contamination. The clean-target dataset we use is CIFAR-10 \citep{Krizhevsky2009LearningML} and we consider several contamination mechanisms. We evaluate generation quality using the Fréchet Inception distance (FID) \citep{NIPS2017_8a1d6947}.

\citet{li2026robustdiffusionmodelsdivergenceinduced} perform similar studies on CIFAR-10 as we do and we adopt several contamination mechanisms used in their experiments. They use a different framework and report that they deliberately use a small model and limited training budget. We use a well-performing Flow Matching configuration instead and first verify that both regression objectives attain FID values comparable to the established result on clean CIFAR-10. This shows that the $L_{2,1}$ objective achieves generation quality comparable to standard Flow Matching on clean data and allows us to study its robustness in a competitive Flow Matching setting. Because the modeling frameworks, training budgets, and evaluation procedures differ, the absolute FID values across the two studies are not directly comparable.
A limitation of our results is that the theoretical results concern optimal vector fields at the population level. They do not account for estimation from finite data, approximation by neural networks, or optimization error. The extent to which the theoretical robustness properties carry over to trained models therefore remains unresolved.
To perform the experiments, we use the U-Net architecture from \citet{tong2024improving,tong2023simulation}\footnote{
\url{https://github.com/atong01/conditional-flow-matching}}. The difference between our method and classic flow matching is the norm training objective: squared Euclidean vs. $L_{2,1}$. All remaining components of training and sampling are held fixed across experiments. \citet{tong2024improving} report a FID of $\num{3.659}$ for independent FM on CIFAR-10. We use a standard Gaussian source distribution, linear interpolation and independent couplings. We use the same interpolation, optimizer, batch size and training budget. Specifically, models are trained for $\num{400000}$ gradient steps with Adam and a learning rate of $2\times10^{-4}$, batch size of $\num{128}$, $\num{5000}$ warm-up steps, gradient clipping at $1.0$, and an EMA decay of $0.9999$. We report the mean and standard deviation over three seeds. We fix the seeds $1,2,3$. Both methods use the same initialization seeds, and contamination is applied to the same images. FID is evaluated against the clean CIFAR-10 training set using $\num{50000}$ samples generated from the exponential moving average (EMA) model using the adaptive Dormand-Prince (DOPRI5) \citep{DORMAND198019} solver with tolerance $10^{-5}$.
\subsection{Contamination mechanisms}
As discussed in Section \ref{sec:rob_ana} we use the contamination model
\begin{equation*}
    P^{\textup{cont}} = (1-\epsilon)P + \epsilon Q
\end{equation*}
where $P$ denotes the CIFAR-10 distribution. We consider contamination proportions $\epsilon \in \{0.05,0.1,0.15\}$ and one clean reference setting where $\epsilon=0$. 
The contamination models we are testing are described in Table \ref{tab:contaminations} and an example visualization is included in the supplementary material in Figure \ref{fig:contamination_examples}.
Theorem \ref{thm:bd} suggests that our Flow Matching method should be particularly advantageous when the contaminating distribution is well separated from the clean target. Motivated by this observation, we consider contamination mechanisms with different degrees of severity and structure. These range from replacement by noise through global and localized pixel perturbations, to blur transformations. The  contaminating distribution $Q$ may either be a completely independent replacement distribution or induced by applying corruption to clean images. Several of our contamination models are taken from \cite{li2026robustdiffusionmodelsdivergenceinduced}. Note that they combine the contamination models into one mixture contamination distribution, while we evaluate them separately.

\begin{table}[t]\label{table_conts}
    \centering
    \small
    \caption{Contamination mechanisms used in the experiments. Images
    represented in $[-1,1]^{3\times32\times32}$.}
    \label{tab:contaminations}
    \begin{tabularx}{\linewidth}{@{}lXl@{}}
        \toprule
        Contamination & Description & Parameters \\
        \midrule
        Random noise
        & Replace the entire image by independent uniform noise
        & $U[-1,1]$ \\

        Gaussian noise
        & Add independent Gaussian noise and clip to $[-1,1]$
        & $\sigma=0.15$ \\

        Salt and pepper
        & Replace selected pixels by either $-1$ or $1$
        & $\rho=0.05$ \\

        Patch noise
        & Replace a uniformly located square patch by uniform noise
        & $8\times8$ patch \\

        Inversion
        & Invert all normalized pixel values, $T(x)=-x$
        & -- \\

        Gaussian blur
        & Convolve the image with an isotropic Gaussian kernel
        & $\sigma=1.5$, $11\times11$ \\

        Motion blur
        & Convolve the image with a horizontal motion-blur kernel
        & kernel size $7$ \\
        \bottomrule
    \end{tabularx}
\end{table}


\subsection{Results}
\begin{table}[t]
    \centering
    \small
    \caption{FID on clean CIFAR-10 (lower is better). Entries report the
    mean $\pm$ standard deviation over three seeds. The lower mean is shown
    in bold.}
    \label{tab:fid_clean}
    \begin{tabular}{@{}lc@{}}
        \toprule
        Objective & FID \\
        \midrule
        Squared   & $3.70 \pm 0.08$ \\
        Unsquared & $\mathbf{3.50} \pm 0.01$ \\
        \bottomrule
    \end{tabular}
\end{table}

\begin{table}[t]
    \centering
    \small
    \setlength{\tabcolsep}{5pt}
    \caption{FID on clean CIFAR-10 (lower is better) after training under
    contamination. Entries report the mean $\pm$ standard deviation over
    three seeds. The lower mean within each pair is bold.}
    \label{tab:fid_contaminated}
    \begin{tabular}{@{}llccc@{}}
        \toprule
        Contamination & Objective
        & $\epsilon=0.05$
        & $\epsilon=0.10$
        & $\epsilon=0.15$ \\
        \midrule
        Random noise
        & Squared   & $9.69 \pm 0.20$          & $19.16 \pm 0.33$          & $29.73 \pm 0.61$ \\
        & Unsquared & $\mathbf{6.82} \pm 0.26$ & $\mathbf{12.21} \pm 0.28$ & $\mathbf{19.40} \pm 0.82$ \\
        \addlinespace[2pt]

        Gaussian noise
        & Squared   & $3.79 \pm 0.06$          & $4.92 \pm 0.17$          & $6.72 \pm 0.20$ \\
        & Unsquared & $\mathbf{3.51} \pm 0.07$ & $\mathbf{4.17} \pm 0.06$ & $\mathbf{5.31} \pm 0.05$ \\
        \addlinespace[2pt]

        Salt and pepper
        & Squared   & $4.45 \pm 0.04$          & $6.49 \pm 0.17$          & $9.44 \pm 0.53$ \\
        & Unsquared & $\mathbf{3.92} \pm 0.10$ & $\mathbf{5.00} \pm 0.09$ & $\mathbf{6.65} \pm 0.32$ \\
        \addlinespace[2pt]

        Patch noise
        & Squared   & $3.98 \pm 0.05$          & $4.61 \pm 0.07$          & $5.49 \pm 0.08$ \\
        & Unsquared & $\mathbf{3.68} \pm 0.07$ & $\mathbf{4.10} \pm 0.08$ & $\mathbf{4.70} \pm 0.11$ \\
        \addlinespace[2pt]

        Inversion
        & Squared   & $3.96 \pm 0.10$          & $4.84 \pm 0.18$          & $6.06 \pm 0.15$ \\
        & Unsquared & $\mathbf{3.66} \pm 0.07$ & $\mathbf{4.37} \pm 0.06$ & $\mathbf{5.52} \pm 0.13$ \\
        \addlinespace[2pt]

        Gaussian blur
        & Squared   & $\mathbf{4.58} \pm 0.07$ & $\mathbf{6.18} \pm 0.10$ & $\mathbf{8.84} \pm 0.25$ \\
        & Unsquared & $5.09 \pm 0.11$          & $7.98 \pm 0.09$ & $11.85 \pm 0.07$ \\
        \addlinespace[2pt]

        Motion blur
        & Squared   & $\mathbf{4.45} \pm 0.06$ & $\mathbf{6.13} \pm 0.11$ & $\mathbf{8.48} \pm 0.13$ \\
        & Unsquared & $4.71 \pm 0.09$          & $7.18 \pm 0.22$          & $10.40 \pm 0.21$ \\
        \bottomrule
    \end{tabular}
\end{table}

The results for the clean experiments are presented in Table \ref{tab:fid_clean}. Both the standard and our method achieve similarly low FID values, while our method yields a lower mean FID. This is close to the $3.659$ baseline reported by \citet{tong2024improving}.

Table \ref{tab:fid_contaminated} reports the FID for the contaminated experiments with $\epsilon \in \{ 0.05,0.1,0.15\}$. Our method performs better, across all proportions of contamination, for Random noise, Gaussian noise, Salt and Pepper, Patch noise and Inversion. The standard method performs better, again across all proportions of contamination, for the Gaussian blur and Motion blur. 
\subsection{Discussion and Limitations}
The results on clean data suggest that in this particular model, the modified Flow Matching objective can preserve generation quality. Lemma \ref{lem:symmetry} gives a sufficient condition for this to hold. However, we cannot verify its assumptions for the CIFAR-10 distribution, although the empirical observations are consistent with it. 

The pattern for the contaminated models suggests that the $L_{2,1}$-Flow Matching method is effective for certain kinds of contamination. In particular, in the models under which the $L_{2,1}$ method performs better, contamination consists of strong perturbations of pixels, individually or across the whole image. We suspect that these contamination models result in substantial deviations from the clean image distribution, under which the $L_{2,1}$ objective is beneficial. The standard method performs better for both Gaussian blur and Motion blur. A possible explanation is that the blurring we applied produces a contamination that is less separated from the original distribution. This hypothesis is partially supported by Theorem \ref{thm:bd}. We showed that under the stated assumptions, for bounded clean data with increasingly distant Dirac contamination, eventually the $L_{2,1}$ method will yield a better result, in the Wasserstein metric. Moreover, this resembles the intuition that the geometric median is generally more useful for contamination that is strongly separated from the original data. 
Nevertheless, we should be careful with this intuition. The intuition regarding the geometric median applies only partially, because the velocity distributions $\textup{Law}(X_1 -X_0 \vert X_t=x)$, which are the basis of the geometric median, have different mixture weights than the contaminated target distributions.
\section{Conclusion}
In this work we proposed an adaptation of the Flow Matching method, obtained by replacing the regression objective with an $L_{2,1}$ objective. We gave a characterization of the solution as a conditional geometric median and established conditions under which this changed objective yields the same result as the standard method. We then investigated the effect that contamination has on the vector fields and provided bounds on the influence of contamination. We have established improvement in the Wasserstein distance for our method in specific contamination settings under stated assumptions. Empirically, the method achieves comparable performance on clean CIFAR-10 and improves generation quality under several, though not all, contamination mechanisms. We do not establish general regularity or uniqueness results for trajectories induced by the $L_{2,1}$ formulation. Moreover, pointwise robustness of the velocity field does not by itself guarantee that the generated distribution is closer to the clean target. Extending our distributional guarantees beyond the specific contamination settings that we analyzed therefore remains an important direction for future research.

\subsection*{Acknowledgments}
We acknowledge funding by the German Research Foundation under the number OC 150/7-1.

\subsection*{AI use statement}

In this work, we used generative AI tools to assist with formulating mathematical claims, providing critical ingredients for proving mathematical claims, assisting in writing of proofs and implementing methods and proposing or refining hypotheses.
We have not used generative AI tools for generating synthetic data sets, helping develop theoretical models or conceptual frameworks, designing or providing feedback on research methodology or experiments, cleaning and reformatting datasets, supporting qualitative and thematic data analysis, or to help interpret results.  Assisting with translation is not applicable to this work.
Additionally, we used generative AI tools for creating or modifying scientific figures or images, creating or editing software code, summarizing or analyzing existing literature, sourcing/searching for information, editing a research paper to improve readability, identifying relevant literature, suggesting a structure for a research paper. 

We have reviewed all AI-assisted work. In particular, we independently checked all AI-assisted mathematical claims and proof ideas. Moreover, we inspected the reviewed AI-assisted code and verified literature-related suggestions against their original sources. We take responsibility for the final content of this work, including text, claims or artifacts produced with the aid of generative AI.

\begin{appendices}
    
\newpage
\section{Proofs}\label{appen:proofs}
Here we prove the statements from the main text. Throughout this appendix, we retain the notation and assumptions introduced in the main text. For convenience, we reintroduce the statements we prove. 

\begin{lemma*}[\ref{lemma:field_gmed}]
 Let $\fielduns$ be a measurable minimizer of \eqref{unsquared_reg}. Then it holds that
    \begin{equation*}
        \fielduns \in \textup{Gmed}(X_1-X_0 \vert X_t = x),
    \end{equation*}
    for almost every $t$ and $P_t$-almost every $x$. Moreover, if this conditional distribution is not concentrated on an affine line for almost every $t$ and $P_t$-almost every $x$, then the minimizer is unique up to almost everywhere equality. 
\end{lemma*}
\begin{proof}
     Let $K_t(x ,\cdot)$ denote a jointly measurable probability kernel representing the conditional law of $X_1 -X_0 $ given $X_t=x$. Then
     \begin{align*}
         \mathcal{L}(v) &= \int_0^1\mathbb{E}[ \lVert v_t(X_t) - (X_1-X_0) \rVert ] \textup{d}t \\
         &= \int_0^1 \int_{\mathbb{R}^n \times \mathbb{R}^n} \lVert v_t(x) - z \rVert K_t(x, \textup{dz}) P_t(\textup{d}x) \textup{d}t.
     \end{align*}
     For fixed $t,x$ the inner integral 
     \begin{equation*}
         r(t,x,a) \coloneqq\int_{\mathbb{R}^n} \lVert a-z \rVert K_t(x ,\textup{d}z)
     \end{equation*}
     is minimized exactly when $a$ is a geometric median of $K_t(x,\cdot)$. This holds for almost every $t,x$. Assume that $v_t^1(x)$ would not minimize $r(t,x,\cdot)$ on a set that has positive measure that we denote by $E$. By continuity in the last argument and since $\mathbb{Q}^{n}$ is dense there has to exist a rational vector $q^\ast \in \mathbb{Q}^n$ such that
     \begin{equation*}
         r(t,x,q^\ast) < r(t,x,v_t^1(x))
     \end{equation*}
     for $(t,x)$ in a set of positive measure. To see that, let 
     \begin{equation*}
         A_q \coloneqq \{ (t,x): r(t,x,q) <r(t,x,v_t^{1}(x)) \}.
     \end{equation*}
     Since $r$ is continuous in its last argument and $\mathbb{Q}^n$ is dense we can write
     \begin{equation*}
         E = \bigcup_{q \in \mathbb{Q}^n} A_q.
     \end{equation*}
Since $E$ has positive measure, there has to exist a $q^\ast$ such that $A_{q^\ast}$ has positive measure. Consequently, replacing $v_t^1(x)$ with $q^\ast$ on $A_{q^\ast}$ would decrease $\mathcal{L}$, contradicting minimality.
     The uniqueness claim follows immediately from \citet{10.1214/aos/1176350511}.
   \end{proof}

\begin{lemma*}[\ref{lem:symmetry}]
Assume that, for almost every $t$ and $P_t$-almost every $x$,
the conditional distribution of $X_1-X_0$ given $X_t=x$ is not concentrated on an affine line and is centrally symmetric around its mean, i.e., for any random variable having this distribution, say Y, we have that  $ Y-\mathbb{E}[Y]$ matches, in distribution, $-(Y-\mathbb{E}[Y])$. Then we have that
        \begin{equation*}
            \fielduns = \fieldsquared
        \end{equation*}
        for almost every $t$ and $P_t$-almost every $x$.
\end{lemma*}
\begin{proof}
This holds for general distributions. Let $\mu$ be a distribution satisfying the assumptions above. Let
\begin{equation*}
    m = \int x \mu(\textup{d}x)
\end{equation*}
be the expectation of $\mu$. Consider 
\begin{equation*}
    F(y) = \int \lVert y -x \rVert \mu (\textup{d}x).
\end{equation*}
Since $\mu$ has finite expectation, the minimizer of $F$ is the geometric median, which is unique by assumption. Since $\mu$ is assumed to be centrally symmetric around its mean, it holds that
\begin{equation*}
    F(y) = F(2m-y).
\end{equation*}
Since $F $ is convex
\begin{equation*}
    F(m) = F\left( \frac{y+(2m-y)}{2} \right) \leq \frac{F(y) + F(2m-y)}{2} = F(y),
\end{equation*}
implying that $m$ is the unique geometric median. Applying this result to the conditional distribution of
$X_1-X_0$ given $X_t=x$ and using the characterizations of
$\fielduns$ and $\fieldsquared$ proves the claim.
\end{proof}

\begin{corollary*}[\ref{corr:gauss:equality}]
    If the source distribution is a standard Gaussian distribution $X_0 \sim \mathcal{N}(0, \idmat)$ and the target distribution is nondegenerate Gaussian $X_1 \sim \mathcal{N}(\mu, \Sigma)$ where $X_0$ and $X_1$ are independent, then \begin{equation*}
            \fielduns = \fieldsquared
        \end{equation*}
        for almost every $t$ and $P_t$-almost every $x$.
\end{corollary*}
\begin{proof}
  Since $X_0$ and $X_1$ are independent Gaussian random variables it follows that $X_t = (1-t)X_0 +tX_1$ is Gaussian. As both $X_1 - X_0$ and $X_t$ are linear combinations of $X_0 $ and $X_1$, the pair $(X_1-X_0 , X_t)$ is jointly Gaussian. Thus $X_1 -X_0$ given $X_t = x$ is nondegenerate Gaussian. The statement then follows since Gaussian distributions have a unique geometric median equal to their mean. 
\end{proof}

\begin{theorem*}[\ref{thm:cond_vel}]
Let $P_t$ be the law of $X_t$. Under the standing assumptions, we have that:
    \begin{equation*}
        \int_0^1 \int_{\mathbb{R}^n} \lVert \fielduns - \fieldsquared \rVert^2 P_t(\textup{d}x) \textup{d}t \leq \int_0^1 \mathbb{E}\lVert (X_1-X_0) - v_t^{2} (X_t) \rVert^2 \textup{d}t
    \end{equation*}
    where the right-hand side is the optimal squared FM regression loss. 
\end{theorem*}

\begin{proof}
    The following inequalities hold for almost every $t$ and $P_t$-almost every $x$.
    Let 
\begin{equation*}
    \fieldsquared = \mathbb{E}[X_1 - X_0 \vert X_t = x]
\end{equation*}
and 
\begin{equation*}
    \fielduns \in \textup{Gmed}[X_1 - X_0 \vert X_t = x].
\end{equation*}
Then 
\begin{equation*}
    \lVert \fieldsquared - \fielduns \rVert = \lVert  \mathbb{E}[(X_1 - X_0) - \fielduns \vert X_t = x]  \rVert \leq  \mathbb{E}[\lVert (X_1 - X_0) - \fielduns \rVert \vert X_t = x]. 
\end{equation*}
Since $\fielduns$ is a geometric median of the conditional
distribution, it minimizes the conditional expected Euclidean
distance. Consequently
\begin{equation*}
     \lVert \fieldsquared - \fielduns \rVert \leq \mathbb{E}[\lVert (X_1 - X_0) - \fielduns \rVert \vert X_t = x] \leq \mathbb{E}[\lVert (X_1 - X_0) - \fieldsquared \rVert \vert X_t = x].
\end{equation*}
Moreover
\begin{align*}
    \mathbb{E}[\lVert (X_1 - X_0) - \fieldsquared \rVert \vert X_t = x] &\leq \sqrt{\mathbb{E}[\lVert (X_1 - X_0) - \fieldsquared \rVert^2 \vert X_t = x] }.
\end{align*}
    Consequently 
    \begin{equation*}
         \lVert \fieldsquared - \fielduns \rVert^2 \leq \mathbb{E}[\lVert (X_1 - X_0) - \fieldsquared \rVert^2 \vert X_t = x] 
    \end{equation*}
    and integrating with respect to $P_t$ yields
    \begin{align*}
         \int_0^1 \int_{\mathbb{R}^n} \lVert \fielduns - \fieldsquared \rVert^2 P_t(\textup{d}x) \textup{d}t \leq \int_0^1 \mathbb{E} \lVert (X_1-X_0) - v^2_t(X_t) \rVert^2 \textup{d}t
    \end{align*}
\end{proof}

\begin{lemma*}[\ref{unsq_cont_bound}]
For almost every $t$ and $P_t^\textup{cont} $-almost every $x$ satisfying  $\gamma_t(x) < 1/2$, there exists $C_t(x) \ge 0$, depending only on the clean model and not on the contamination, such that
\begin{equation*}
     \lVert \fieldunscont - \fielduns \rVert \leq \frac{2(1-\gamma_t(x))}{1-2\gamma_t(x)} C_t(x).
\end{equation*}    
\end{lemma*}

\begin{proof}
Fix $t$ and $x$. We have shown in \eqref{eq:prob_decom} that we can decompose the velocity distribution as 
\begin{equation*}
    P_{t,x}^{\textup{cont}} = (1-\gamma_t(x))P_{t,x} + \gamma_t(x) Q_{t,x}. 
\end{equation*}
Let $F(z) = \int \lVert y-z\rVert P_{t,x} (\textup{d}y)$ and $G(z)=\int \lVert y-z\rVert Q_{t,x} (\textup{d}y)$. By definition $\fielduns$ minimizes $F$, because $P_{t,x}$ is the clean conditional velocity distribution. On the other hand, by the decomposition of $P_{t,x}^{\textup{cont}}$, it holds that $\fieldunscont $ minimizes 
\begin{equation*}
    F_{\textup{cont}}(z) = (1-\gamma_t(x))F(z) + \gamma_t(x) G(z).
\end{equation*}
Denote $\fielduns = m$ and $\fieldunscont = m^{\ast}$. Due to the optimality mentioned above
\begin{equation}\label{opt_ineq}
    (1-\gamma_t(x))F(m^\ast) + \gamma_t(x) G(m^\ast) \leq (1-\gamma_t(x))F(m) + \gamma_t(x) G(m).
\end{equation}
Let $d = \lVert m^\ast - m \rVert $. 

Using the reverse triangle inequality we have
\begin{equation*}
    \lVert y - m^\ast \rVert \geq \lVert m^\ast - m \rVert - \lVert y-m \rVert
\end{equation*}
and 
\begin{equation*}
    \lVert y - m^\ast \rVert \geq  \lVert y-m \rVert - \lVert m^\ast - m \rVert.
\end{equation*}
By taking the appropriate expectations this implies 
\begin{equation*}
    F(m^\ast) \geq d - F(m)
\end{equation*}
and 
\begin{equation*}
    G(m^\ast) \geq G(m) -d.
\end{equation*}
Plugging those into \eqref{opt_ineq}:
\begin{equation*}
    (1-\gamma_t(x))(d-F(m)) + \gamma_t(x)(G(m)-d) \leq (1-\gamma_t(x)) F(m) + \gamma_t(x) G(m)
\end{equation*}
after reformulating 
\begin{equation*}
    (1-2\gamma_t(x))d - (1-\gamma_t(x))F(m) + \gamma_t(x)G(m) \leq (1-\gamma_t(x)) F(m) + \gamma_t(x)G(m)
\end{equation*}
equivalent to 
\begin{equation*}
    (1-2\gamma_t(x))  d \leq 2(1-\gamma_t(x))F(m).
\end{equation*}
Since by assumption $\gamma_t(x) < 1/2$ it holds that $1-2\gamma_t(x) > 0$ and we conclude
\begin{equation*}
    d \leq \frac{2(1-\gamma_t(x))}{1-2\gamma_t(x)}F(m).
\end{equation*}
With 
\begin{equation*}
    C_t(x)  = \int \lVert y- v_t^1(x)\rVert P_{t,x}(\textup{d}y)
\end{equation*}
the claim follows.
\end{proof}

\begin{theorem*}[\ref{thm:dirac}]
Let $\delta_0 $ be a Dirac distribution in $\mathbb{R}^n$ centered at $0$. Let $0 < \epsilon<1/2$. For $b \in \mathbb{R}^n \setminus \{0\}$, consider the contaminated distribution
\begin{equation*}
    P^{\textup{cont}} = (1-\epsilon)\delta_0 + \epsilon \delta_b.
\end{equation*}
For every pointwise $L_{2,1}$-velocity field $v^{1,P^{\textup{cont}}}$ coming from a Borel geometric-median selection as defined in \eqref{concrete:def:field}, the associated ODE admits a solution process $(Z_t)_{t\in[0,1)} $ with $Z_0 \sim \mathcal{N}(0,\idmat_n)$. Every such solution converges almost surely as $t \nearrow 1 $ and its terminal law $\rho_1 = \textup{Law}(\lim_{t \nearrow1} Z_t)$ satisfies
    \begin{equation*}
        W_1(\rho_1,\delta_0)  < W_1(P^{\textup{cont}},\delta_0).
    \end{equation*}
\end{theorem*}

\begin{proof}
As described in the main text, we will condition on the endpoint being either the clean or the contaminated target. We consider $0 \leq t < 1$. Conditioned on either the clean or the contaminated endpoint, the distribution of the interpolation is Gaussian
\begin{equation}\label{interpolationDist}
    X_t \vert (X_1=\xi) \sim \mathcal{N}(t \xi , (1-t)^2 \idmat)
\end{equation}
where $\xi \in \{ 0 , b \}$. Define the posterior contamination probability $ \gamma_t(x) = \mathbb{P}[X_1 = b \vert X_t =x ]$ and 
\begin{align*}
    \posteriorQuot = \frac{\gamma_t(x)}{(1-\gamma_t(x))} 
    = \frac{\mathbb{P}[X_1 = b \vert X_t =x ]}{\mathbb{P}[X_1 = 0 \vert X_t =x ]}.
\end{align*}
We want to analyze this quantity, since it explains the behavior of the $L_{2,1}$ field. From \eqref{interpolationDist} we obtain that $X_t \vert (X_1= \xi)$ admits a density. Using $\mathbb{P}[X_1 = b] = \epsilon$ and the law of total probability we obtain that for the density $p_t(x) $ of $X_t$ the following holds:
\begin{equation*}
    p_t(x) = (1-\epsilon) p_t(x \vert X_1 = 0) + \epsilon p_t(x \vert X_1 = b).
\end{equation*}
Consequently using Bayes' theorem
\begin{equation*}
    \gamma_t(x) = \frac{\epsilon p_t(x \vert X_1 = b)}{(1-\epsilon) p_t(x \vert X_1 = 0) + \epsilon p_t(x \vert X_1 = b)}.
\end{equation*}
Similarly, 
\begin{equation*}
   1- \gamma_t(x) = \frac{(1-\epsilon) p_t(x \vert X_1 = 0)}{(1-\epsilon) p_t(x \vert X_1 = 0) + \epsilon p_t(x \vert X_1 = b)}.
\end{equation*}
For $\posteriorQuot$ the common denominator cancels and it follows that
\begin{equation*}
    \posteriorQuot = \frac{\epsilon}{1-\epsilon} \frac{p_t(x \vert X_1 = b)}{p_t(x \vert X_1 = 0)}. 
\end{equation*}
With the characterization of these densities in \eqref{interpolationDist} we obtain
\begin{equation*}
    \posteriorQuot = \frac{\epsilon}{1-\epsilon} \exp \left(-\frac{\lVert x- tb \rVert^2 - \lVert x\rVert^2 }{2(1-t)^2} \right).
\end{equation*}
Taking the logarithm yields
\begin{equation}\label{log_theorem}
    \log \posteriorQuot = \log \frac{\epsilon}{1-\epsilon} + \frac{t \langle x,b \rangle }{(1-t)^2} - \frac{t^2 \lVert b \rVert^2}{2(1-t)^2}.
\end{equation}

Conditioned on an endpoint and the interpolation point, the source point is deterministic. Conditioning on $X_t = x$ and $X_1 = \xi$ gives
\begin{equation*}
    X_0 = \frac{x - t\xi}{1-t}, \quad X_1 -X_0 = \xi - X_0 = \frac{\xi-x}{1-t}.
\end{equation*}
For each $\xi \in\{ 0,b\}$ define the conditional field 
\begin{equation*}
    u_t^\xi(x) = \frac{\xi-x}{1-t}.
\end{equation*}
In the sense of probability laws we have 
\begin{equation*}
    \textup{Law}(X_1 - X_0 \vert X_t = x) = (1-\gamma_t(x)) \delta_{u^{0}_t(x)} + \gamma_t(x) \delta_{u^{b}_t(x)}.
\end{equation*}
 For fixed \(t<1\) and \(x\), this conditional law is a two-point mixture. If \(\gamma_t(x)\neq1/2\), its geometric median is the atom with higher assigned probability. 
Note that $\gamma_t(x)< 1/2 \iff \posteriorQuot < 1 \iff \log \Gamma_t(x) < 0$. Fix any Borel geometric-median selection $m^{P^{\textup{cont}}}$ as defined in the main text and write 
\begin{equation*}
    \fielduns = v_t^{1,P^\textup{cont}}(x) = \frac{m_t^{P^{\textup{cont}}}(x) -x}{1-t}
\end{equation*}
Since the relevant conditional law is a two-point mixture as discussed, the field satisfies
\begin{equation*}
    \fielduns= \begin{cases}
            u^{0}_t(x) & \text{if $\posteriorQuot < 1$} \\
            (1-\lambda_t(x))u^{0}_t(x) + \lambda_t(x)   u^{b}_t(x)         & \text{if $\posteriorQuot = 1$}  \\
            u^{b}_t(x) & \text{if $\posteriorQuot > 1$},
        \end{cases}
\end{equation*}
where $\lambda_t(x) \in [0,1]$ is determined by the fixed Borel selection, which we leave arbitrary. Note that this is relevant since for $\gamma_t(x) = 1/2$ both mixture weights are equal, thus the set of geometric medians is the line segment $[u^{0}_t(x),u^{b}_t(x)]$. For fixed source points, we can analyze the trajectories induced by the field.
For a starting point $x_0$, the trajectory induced by the uncontaminated field is $T_t(x_0)=(1-t)x_0 $. Let $c = -\log \frac{\epsilon}{1-\epsilon} >0$ and $S(x_0) = \frac{\langle b,x_0\rangle}{\lVert b \rVert}$. Using \eqref{log_theorem}:
\begin{equation*}
    \log \Gamma_t(T_t(x_0)) = -c +\frac{t}{1-t} \lVert b \rVert S(x_0) - \frac{1}{2} \frac{t^2}{(1-t)^2} \lVert b \rVert ^2.
\end{equation*}
Let $z = \frac{t}{(1-t)}$. Then this is a concave quadratic in $z$:
\begin{equation*}
    \log \Gamma_t(T_t(x_0)) = -c +  \lVert b \rVert S(x_0) z- \frac{1}{2}\lVert b \rVert ^2 z^2 \eqqcolon G_{x_0}(z).
\end{equation*}
Since the transformation $ t \mapsto z$ is a continuous increasing function from $[0,1)$ onto $[0,\infty)$ it suffices to study the time-reparametrized quadratic $G_{x_0}$. If $S(x_0) > 0$ then its maximum over $z\ge0$ is $-c+\frac{1}{2}S(x_0)^2$, while for $S(x_0) \leq 0$ its maximum is $-c <0$. Consequently, if
\begin{equation*}
    S(x_0) < \sqrt{2c},
\end{equation*}
then
\begin{equation*}
     \log \Gamma_t(T_t(x_0)) < 0,
\end{equation*}
for every $t<1$. We are interested in the sign of $G_{x_0}$, since 
\begin{equation*}
    G_{x_0}(z) < 0 \iff \Gamma_{\frac{z}{1+z}}(T_{\frac{z}{1+z}}(x_0)) = \Gamma_t(T_t(x_0)) < 1.
\end{equation*}
Define $H_{\textup{clean}} = \{ x_0 \in \mathbb{R}^n: S(x_0) < \sqrt{2c}\}$ and let $x_0 \in H_\textup{clean}$. By what we have shown above, $G_{x_0}(z) < 0$ for every $z \geq 0$. In particular this implies that the vector field selects the clean velocity along the candidate clean trajectory
\begin{equation*}
    v_t^{1}(T_t(x_0)) = u_t^0(T_t(x_0)) =\frac{-T_t(x_0)}{1-t}.
\end{equation*}
Since
\begin{equation*}
    \frac{\textup{d}}{\textup{d}t} T_t(x_0) = -x_0 = \frac{-T_t(x_0)}{1-t}
\end{equation*}
the clean trajectory is indeed a trajectory of the $L_{2,1}$-field ODE on $[0,1)$.
Note that $\Gamma$ is continuous and the trajectory $T_t(x_0)$ is also continuous. Since $\Gamma_t(T_t(x_0)) <1$, for every $t<1$ there exists a neighborhood around $(t,T_t(x_0))$ such that $\Gamma <1$. Consequently the geometric median field agrees with 
\begin{equation*}
    u_s^0 (y) = -\frac{y}{1-s} 
\end{equation*}
for all $s,y$ in that neighborhood. This field is locally Lipschitz in space and hence the clean trajectory is indeed the unique trajectory starting from $x_0$.  

Hence, starting points $x_0 \in H_\textup{clean}$ follow $T_t(x_0)$ and consequently get transported to the clean target since $T_t(x_0) = (1-t)x_0 \to 0$ for $t \nearrow 1$.

Now define $H_{\textup{cont}}  = \{ x_0 \in \mathbb{R}^n: S(x_0) > \sqrt{2c}\}$ and let $x_0 \in H_{\textup{cont}}$. Consider $G_{x_0}(z) $ as defined above. Since $S(x_0) > \sqrt{2c}$ it has a positive maximum and consequently has the two positive roots
\begin{equation*}
    z_{\pm} = \frac{S(x_0) \pm \sqrt{S(x_0)^2 -2c}}{\lVert b \rVert}.
\end{equation*}
Consider $ z_- = \frac{S(x_0) - \sqrt{S(x_0)^2 -2c}}{\lVert b \rVert} $ and $t_- = \frac{z_-}{1+z_-}$. Consider any locally absolutely continuous trajectory with initial value $x_0$, solving the flow ODE for almost every $t$. Analogously to the previous argument, since $G_{x_0}(z) <0$ for $0 \leq z <z_-$ the trajectory initially follows the clean field for $0 \leq t < t_-$. After a potential switch, the candidate trajectory is $T_t(x_0) = (1-t)Y(z)$ where $Y(z)= x_0 + (z-z_-)b$. Its posterior log-odds are
\begin{equation*}
    L_{x_0}(z) = -c + \langle Y(z),b \rangle z- \frac{1}{2} \lVert b \rVert ^2 z^2
\end{equation*}
with $L_{x_0}(z_-)=0$ and since $Y'(z) = b$ we have 
\begin{equation*}
    L'_{x_0}(z) = \langle Y(z),b \rangle = \lVert b \rVert S(x_0) + (z-z_-) \lVert b \rVert ^2 > 0 \quad \text{for every } z > z_-,
\end{equation*}
implying $L_{x_0}(z) > 0$ for every $z > z_-$. Note that $ G^\prime_{x_0}(z_-) > 0 $ and $ L^\prime_{x_0}(z_-) > 0 $. 
Since $dz/dt=(1-t)^{-2}>0$, the corresponding time derivatives are also positive at the switching point.
By a continuity argument analogous to the one above,
the derivatives of the log-odds along both branch velocities
remain bounded below by a common positive constant in a
neighborhood of the switching point. On the tie surface, the velocity is a convex combination of 
$u_t^0(x)$ and $u_t^b(x)$. Write $l(t,x) = \log \Gamma_t(x)$. For the selected velocity, the chain rule gives
\begin{equation*}
    \partial_t l + \langle \nabla_x l , v_t^1 \rangle = (1-\lambda ) ( \partial_t l+ \langle \nabla_x l ,u_t^0 \ \rangle) + \lambda ( \partial_t l + \langle \nabla_x l ,u_t^b \ \rangle) \geq \eta > 0,
\end{equation*}
where $\lambda \in [0,1]$ is the coefficient of the convex combination and $\eta$ is the common positive lower bound. All quantities are evaluated at $(t,x)$.  
 Hence, along any
absolutely continuous solution through this point, the
log-odds increase strictly for a sufficiently short time,
so the solution immediately enters the contaminated region.
Local uniqueness there identifies it with the candidate
trajectory, which remains in that region since
$L_{x_0}(z)>0$ for every $z>z_-$.

Hence, the trajectory after $z_- $ is $T_t(x_0)= (1-t)Y(z) = \frac{x_0+(z-z_-)b}{1+z}$. As $t \nearrow1$, $z\to \infty$ so the trajectory converges to $b$. This means that initial points $x_0 \in H_{\textup{cont}}$ get transported to $b$.

Putting this together, the trajectories follow
\begin{equation*}
    T_t(x_0) =  \begin{cases}
            (1-t)x_0 & \text{if $S(x_0) \leq \sqrt{2c}$} \\
            (1-t)x_0 + \max \{0,t-(1-t)z_{-}(x_0) \}b              & \text{if $S(x_0) > \sqrt{2c}$}.
        \end{cases}
\end{equation*}
This map is Borel measurable and the paths are locally absolutely continuous. Let $X_0 \sim \mathcal{N}(0,\idmat_n)$ and define
\begin{equation*}
    Z_t \coloneqq T_t(X_0) , \quad 0 \leq t < 1.
\end{equation*}
This establishes the existence of a solution process.  Our preceding calculations show that for every $x_0$ satisfying $S(x_0) \neq \sqrt{2c}$, $T(x_0)$ satisfies the selected ODE almost everywhere. Since $S(X_0) \sim \mathcal{N}(0,1)$ this condition holds almost surely. Moreover, the preceding uniqueness arguments show that every solution starting from $x_0$ with $S(x_0) \neq \sqrt{2c}$ coincides with $T_t(x_0)$. Consequently, every solution process with the stated Gaussian initial distribution agrees almost surely, for all \(t<1\), with \(T_t(Z_0)\). The previously established trajectory limits therefore give almost-sure convergence and the same terminal law for every such process.

We want to analyze the induced terminal distribution. Since $X_0 \sim \mathcal{N}(0,\idmat)$ it follows that
\begin{equation*}
    S(X_0) = \left \langle \frac{b}{\lVert b \rVert} , X_0  \right \rangle \sim \mathcal{N}(0,1)
\end{equation*} since $S$ is a normalized linear projection. Thus
\begin{equation*}
    \epsilon^\ast = \mathbb{P}[S(X_0) > \sqrt{2c}] = 1- \Phi \left(\sqrt{2c}\right).
\end{equation*}
Consequently, the induced distribution is 
\begin{equation*}
    \rho_1 = (1-\epsilon^\ast) \delta_0 + \epsilon^\ast \delta_b,
\end{equation*}
from which the Wasserstein statement follows immediately since for $a \in [0,1]$:
\begin{equation*}
    W_1 ((1-a)\delta_0 + a \delta_b, \delta_0) = a \lVert b \rVert.
\end{equation*}
Then
\begin{equation*}
    \epsilon^\ast = 1- \Phi \left(\sqrt{2c}\right) 
    < \frac{1}{2} e^{-c} 
    = \frac{\epsilon}{2(1-\epsilon)} < \epsilon,
\end{equation*}
concluding the proof.
\end{proof}

\begin{theorem*}[\ref{thm:bd}]
Let \(P\) be a distribution supported in the closed ball \(\overline{B}_1(0)\). Let \(0<\epsilon<1/2\), and fix a unit vector \(u\in\mathbb R^n\). For \(R>1\), set \(b_R=Ru\) and
$$
P_R^{\mathrm{cont}}
=
(1-\epsilon)P+\epsilon\delta_{b_R}.
$$
For each \(R>1\), let \(v^{1,R}\) be any pointwise \(L_{2,1}\)-velocity field associated with \(P_R^{\mathrm{cont}}\) as defined in \eqref{concrete:def:field}. Assume that its associated ODE admits a solution process with initial distribution \(\mathcal N(0,I_n)\). Let \((Z_t^R)_{t\in[0,1)}\) be any such solution process, and write $\rho_t^R=\operatorname{Law}(Z_t^R).$ Then, for all sufficiently large \(R\),
$$
\limsup_{t\nearrow1}
W_1(\rho_t^R,P)
<
W_1(P_R^{\mathrm{cont}},P).
$$

\end{theorem*}

\begin{proof}

For each $R>1$, let $m^R$ be the Borel geometric-median
selection defining $v^{1,R}$, and let $(Z_t^R)_{t\in[0,1)}$ be any associated
solution process as in the theorem. We suppress the dependence
on $R$ in $m^R$ and $Z_t^R$. By the definition of a solution,
almost surely,
\begin{equation}\label{eq:bounded_integral}
    Z_t = Z_0 + \int_0^t
    \frac{m_s(Z_s)-Z_s}{1-s}\,\mathrm{d}s,
    \qquad \text{for every } 0 \leq t < 1.
\end{equation}

    The beginning of this proof is analogous to the proof of Theorem \ref{thm:dirac}. First we condition on the endpoint. For $0 \leq t < 1$ it holds that
    \begin{equation}
    X_t \vert (X_1=\xi) \sim \mathcal{N}(t \xi , (1-t)^2 \idmat),
    \end{equation}
    where $ \xi \in \overline{B}_1(0) \cup \{ b_R \}$. Since those again admit densities we can write
    \begin{equation*}
        l_\xi (t,x) \coloneqq   \exp \left( \frac{-\lVert x - t \xi \rVert^2}{2(1-t)^2}\right).
    \end{equation*}
    The same Bayes computation as before yields 
    \begin{equation}\label{eq:post:cont:odds}
        \Gamma_t(x) \coloneqq \frac{\gamma_t(x)}{(1-\gamma_t(x))} = \frac{\epsilon}{1-\epsilon} \frac{l_{b_R}(t,x)}{\int l_\xi(t,x) P(\textup{d}\xi)}.
    \end{equation}
    The main difference is that in the denominator we have to integrate over the target distribution, since it is no longer a point mass distribution. The clean posterior is 
    $\nu_{t,x}(\textup{d}y) =  \frac{l_y(t,x) P(\textup{d}y)}{\int l_\xi(t,x) P(\textup{d}\xi)}$. For this distribution it holds that $\textup{supp } \nu_{t,x} \subset \overline{B}_1(0)$. The conditional distribution can now be characterized as
    \begin{equation*}
        \textup{Law}(X_1 \vert X_t=x) = (1-\gamma_t(x)) \nu_{t,x} + \gamma_t(x) \delta_{b_R}.
    \end{equation*}

By the definition of the fixed selection $m$, we therefore have
\begin{equation*}
    m_t(x)
    \in
    \operatorname{Gmed}
    \left(
        (1-\gamma_t(x))\nu_{t,x}
        +\gamma_t(x)\delta_{b_R}
    \right)
\end{equation*}
for every $(t,x)\in[0,1)\times\mathbb{R}^n$.

In the earlier proof, we could give a precise formula for the geometric median, since the velocity distribution consisted of two atoms. Since we are working with a general bounded target distribution, we will instead provide a bound for the geometric median. Fix a threshold
       $ \epsilon < \alpha < 1/2$
    and introduce constants 
    \begin{equation*}
        C_0 = \frac{2(1-\alpha)}{1-2\alpha}, \quad L_0 = \log \left(\frac{\alpha}{1-\alpha} \frac{1-\epsilon}{\epsilon}\right) > 0.
    \end{equation*}
    Along the trajectory define 
    \begin{equation*}
        \tau = \inf \{ t \in [0,1) : \gamma_t(Z_t) \geq \alpha\}
    \end{equation*}
    with the convention that the infimum of the empty set is 1. This is the first time that the posterior contamination probability along the trajectory reaches the threshold. Let \begin{equation*}
        H_R = \{ \tau <1\}
    \end{equation*}
    and $H_R^c$ its complement. 
         On the event $H_R^c$ we have that $\tau = 1$ and thus
    \begin{equation*}
        \gamma_t(Z_t) < \alpha \qquad \text{for every } 0 \leq t < 1. 
    \end{equation*}
    This allows us to bound the median by $C_0$ using Lemma \ref{lemma_moving_median}. Independently of $H_R$, since the clean conditional endpoint distribution is supported in $\overline{B}_1(0)$, the mixed conditional endpoint distribution is supported in $\overline{B}_1(0) \cup \{ b_R\} \subset \overline{B}_R(0)$. Note that the geometric median remains inside $\overline{B}_R(0)$ because otherwise projecting it onto this ball would strictly decrease its distance to every support point. Thus 
    \begin{equation*}
        \lVert m_t(Z_t) \rVert \leq R.
    \end{equation*}
    In particular, this yields the following bound:
      \begin{equation}\label{bd:med_constants}
        \lVert m_r(Z_r) \rVert \leq C_0 \mathbf{1}_{H_R^c} + R \mathbf{1}_{H_R}.
    \end{equation}
    
    Now we want to use the bound on the median to provide a bound on the resulting distribution. Let $\rho_t = \textup{Law}(Z_t)$.
    Take $Y \sim P $ independent of the generated process. Then this yields an admissible coupling between $\rho_t$ and $P$. Since the Wasserstein distance is the infimum of the transport cost of all admissible couplings, we can use this to bound the Wasserstein distance. We have
    \begin{equation}\label{ineq:Wstein}
        W_1(\rho_t , P) \leq \mathbb{E} \lVert Z_t - Y \rVert \leq \mathbb{E}\lVert Z_t\rVert + 1
    \end{equation}
    since $Y \in \overline{B}_1(0)$ almost surely. In the proof of Lemma \ref{lemm:probBound} we have shown 
    \begin{equation}\label{equal:abscont}
        Z_t = (1-t)Z_0 + (1-t) \int_0^t \frac{m_r(Z_r)}{(1-r)^2} \textup{d}r.
    \end{equation}
    Consequently 
    \begin{equation}\label{ineq:expInt}
        \mathbb{E} \lVert Z_t \rVert \leq (1-t) \mathbb{E} \lVert Z_0 \rVert + (1-t) \int_0^t \frac{\mathbb{E} \lVert m_r(Z_r)\rVert}{(1-r)^2} \textup{d}r.
    \end{equation}
    Using \eqref{bd:med_constants}  we obtain
    \begin{equation*}
        \mathbb{E} \lVert m_t(Z_t) \rVert \leq (1- \mathbb{P}(H_R)) C_0 + \mathbb{P}(H_R) R.
    \end{equation*}
    Plugging this into \eqref{ineq:expInt} and explicitly computing the integral yields
    \begin{equation*}
     W_1(\rho_t , P) \leq    \mathbb{E} \lVert Z_t\rVert +1 \leq (1-t) \mathbb{E} \lVert Z_0 \rVert +1+ t ( (1- \mathbb{P}(H_R)) C_0 + \mathbb{P}(H_R) R).
    \end{equation*}
    Taking now $t \nearrow1$ we obtain
    \begin{equation*}
        \limsup_{t \nearrow1} W_1(\rho_t, P) \leq C_0 + 1 + \mathbb{P}(H_R) (R - C_0).
    \end{equation*}
   
    Dividing by R and also sending $R $ to infinity yields
    \begin{align*}
        \limsup_{R \to \infty} \frac{1}{R}\limsup_{t \nearrow1} W_1(\rho_t, P) &\leq \limsup_{R \to \infty} \mathbb{P}(H_R) \\
        &\leq (1-\Phi(\sqrt{2L_0}))
    \end{align*}
    where the last inequality is by Lemma \ref{lemm:probBound}. Since this inequality holds for all $\alpha \in (\epsilon,1/2)$ and the right hand side is decreasing in $\alpha$ we can take the infimum over $\alpha$ and obtain
    \begin{equation*}
         \limsup_{R \to \infty} \frac{1}{R} \limsup_{t \nearrow1} W_1(\rho_t, P) \leq \lim_{\alpha \nearrow 1/2} (1-\Phi(\sqrt{2L_0})) = 1- \Phi\left( \sqrt{2\log\frac{1-\epsilon}{\epsilon}} \right).
    \end{equation*}
    Note that this is the same constant as in the proof of the Dirac contamination theorem. Moreover
    \begin{equation*}
        \epsilon (R -1) \leq  W_1(P_R^{\textup{cont}}, P) \leq \epsilon(R+1),
    \end{equation*}
    implying 
    \begin{equation*}
        \frac{ W_1(P_R^{\textup{cont}}, P)}{R} \to \epsilon.
    \end{equation*}
    Since we have shown that
    \begin{equation*}
         1- \Phi\left( \sqrt{2\log\frac{1-\epsilon}{\epsilon}} \right) < \epsilon
    \end{equation*}
    the statement follows.
\end{proof}

\begin{lemma}\label{lemma_moving_median}
    Consider the setting of Theorem \ref{thm:bd}. Let $ 0 \leq \gamma < 1/2$, let $P $ be any distribution supported in $\overline{B}_1(0)$ and let $m$ be any geometric median of 
    \begin{equation*}
        (1-\gamma)P + \gamma \delta_b
    \end{equation*}
    for $b \in \mathbb{R}^n$. Then
    \begin{equation*}
        \lVert m \rVert \leq \frac{2(1-\gamma)}{1-2\gamma}.
    \end{equation*}
\end{lemma}

\begin{proof}
    Note that this is a similar setup to Lemma \ref{unsq_cont_bound} and a related result is \citep[Lemma 2.1]{Minsker_2015}. Let $F(z) = (1-\gamma) \int \lVert z-y \rVert P(\textup{d}y) + \gamma  \lVert z-b \rVert$. Then $m $ is its minimizer. Consequently $ F(m) \leq F(0)$. Thus 
    \begin{align*}
        F(m) - F(0) &= (1-\gamma) \int ( \lVert m-y \rVert - \lVert y \rVert) P(\textup{d}y) + \gamma( \lVert m-b \rVert - \lVert b \rVert)
    \end{align*}
    implying that the right-hand side is non-positive. Consequently
    \begin{align*}
        0 &\geq  (1-\gamma) \int ( \lVert m-y \rVert - \lVert y \rVert) P(\textup{d}y) + \gamma( \lVert m-b \rVert - \lVert b \rVert) \\
        & \geq (1-\gamma)(\lVert m \rVert -2) - \gamma \lVert m \rVert \\
        &= (1-2\gamma) \lVert m \rVert - 2(1-\gamma).
    \end{align*}
\end{proof}

\begin{lemma}\label{lemm:probBound}
In the setting of Theorem \ref{thm:bd} and with the notation introduced in its proof, 
    for every $\alpha \in (\epsilon,1/2)$,
    \begin{equation*}
        \limsup_{R \to \infty } \mathbb{P}(H_R) \leq (1-\Phi(\sqrt{2L_0}))
    \end{equation*}
    \begin{proof}
       The function $C(\gamma)=2(1-\gamma)/(1-2\gamma)$ is increasing
on $[0,1/2)$. For $t<\tau$, we have $\gamma_t(Z_t)<\alpha$,
so Lemma \ref{lemma_moving_median} yields
\begin{equation*}
    \|m_t(Z_t)\|
    \leq C(\gamma_t(Z_t))
    \leq C(\alpha)=C_0.
\end{equation*} 
By the assumptions of Theorem~\ref{thm:bd}, the sample
paths of $Z$ are locally absolutely continuous and satisfy the
integral equation almost surely. Therefore, for almost every
$t\in[0,1)$,
    \begin{equation*}
    \frac{\mathrm{d}}{\mathrm{d}t}
    \left(\frac{Z_t}{1-t}\right)
    =
    \frac{m_t(Z_t)}{(1-t)^2}.
\end{equation*}
       
       Integrating this from $0 $ to $t$ yields
       \begin{equation*}
           Z_t = (1-t)Z_0 + (1-t) \int_0^t \frac{m_r(Z_r)}{(1-r)^2} \textup{d}r.
       \end{equation*}
       For $0 < t < 1$ with $t \leq \tau$ define
       \begin{equation*}
           c_t = \frac{1-t}{t} \int_0^t \frac{m_r(Z_r)}{(1-r)^2} \textup{d}r.
       \end{equation*}
       This yields 
       \begin{equation}\label{eq:c_cons}
           Z_t = (1-t) Z_0 + t c_t,
       \end{equation}
       and using $\lVert m_t(Z_t) \rVert \leq C_0$ we obtain that $ \lVert c_t \rVert \leq C_0$, which is useful because it is independent of $R$. 
       
       Now suppose $H_R$ occurs. 
       
       The posterior formula is continuous as well as $Z_t$. Consequently, since $\gamma_0(Z_0) =  \epsilon < \alpha$ we have $\gamma_\tau (Z_\tau) = \alpha$. Consider now $t = \tau \, , x = Z_\tau$.    
From the explicit posterior odds formula \eqref{eq:post:cont:odds} this yields 
       \begin{equation*}
           \exp(L_0)  =  \frac{l_{b_R}(t,x)}{\int l_\xi (t,x) P(\textup{d}\xi)} \leq \sup_{\lVert y \rVert \leq 1} \frac{l_{b_R}(t,x)}{l_y(t,x)},
       \end{equation*}
    where \begin{equation*}
        L_0 = \log \left( \frac{\alpha (1-\epsilon)}{(1-\alpha)\epsilon} \right).
    \end{equation*}
       The inequality holds, 
       since $P$ is supported on the unit ball. Denote
       \begin{equation*}
           s = 1-t, z=t/s , S(x_0)= \langle x_0, u \rangle,
       \end{equation*}
      and take $y \in \overline{B}_1(0)$. From \eqref{eq:c_cons} we obtain 
       \begin{equation*}
           x = sZ_0 +  t c_t.
       \end{equation*}
       Denoting $d_y = b_R - y$ we obtain
      \begin{equation*} 
      \log \frac{l_{b_R}(t,x)}{l_y(t,x)}
= z \langle Z_0,d_y\rangle
+ z^2 \langle c_t-y,d_y\rangle
- \frac{z^2}{2}\|d_y\|^2.
      \end{equation*}
      Using $ \lVert y \rVert \leq 1, \lVert c_t \rVert \leq C_0 $ and $b_R = R u$ yields:
      
      \begin{equation*}
         \langle Z_0,d_y\rangle
\leq R S(Z_0)+\|Z_0\|, \, \langle c_t -y , d_y \rangle \leq M_0 (R+1), \,\lVert d_y \rVert \geq R -1.
      \end{equation*}
      
      where $M_0 = C_0 +1$.
      Consequently, for all $ \Vert y \rVert \leq 1$  we have
      \begin{equation*}
         \log \frac{l_{b_R}(t,x)}{l_y(t,x)}
\leq z\bigl(R S(Z_0)+\|Z_0\|\bigr)
- \frac{z^2}{2}A_R.
      \end{equation*}
      where $A_R = (R-1)^2 - 2M_0 (R+1)$.
      Now for all sufficiently large $R$ it holds that $ A_R > 0$. Combining the posterior-odds identity with the preceding uniform estimate yields
    \begin{equation*}
        L_0 \leq zB_R-\frac{z^2}{2}A_R,
    \end{equation*}
    where $B_R = RS(Z_0) + \lVert Z_0 \rVert$. Since $L_0 > 0$ and for $R $ large enough, it holds that $ A_R > 0$, we have $B_R > 0$. Consequently 
    \begin{equation*}
       L_0
\leq \max_{z\geq0}
\left(zB_R-\frac{A_Rz^2}{2}\right)
= \frac{B_R^2}{2A_R}.
    \end{equation*}
    This is equivalent to 
    \begin{equation}\label{maximum_eqv}
      S(Z_0)+\frac{\|Z_0\|}{R}
\geq \frac{\sqrt{2L_0A_R}}{R},
    \end{equation}
    and is now independent of the hitting time $\tau$. 
    Let $G \sim \mathcal{N}(0 ,\idmat_n)$ be a Gaussian random variable independent of the other variables. Let
    \begin{equation*}
        Y = \langle G,u \rangle \sim \mathcal{N}(0,1). 
    \end{equation*}
    
    Since $G$ has the same distribution as $Z_0$ and using \eqref{maximum_eqv}, the following holds:
    \begin{equation}\label{prob_bound_h}
        \mathbb{P}(H_R) \leq \mathbb{P} \left( Y + \frac{\lVert G \rVert}{R} \geq \frac{\sqrt{2 L_0 A_R}}{R} \right).
    \end{equation}
 Note that

    \begin{equation*}
        Y + \frac{\lVert G\rVert}{R} \to Y  , \quad \frac{\sqrt{2 L_0A_R}}{R} \to \sqrt{2L_0},
    \end{equation*}
   almost surely and 
   \begin{equation*}
       \mathbb{P}(Y=\sqrt{2L_0})=0.
   \end{equation*}
   Using the bounded convergence theorem on the indicator variables of the events yields
   \begin{equation*}
       \lim_{R \to \infty} \mathbb{P} \left(Y + \frac{\lVert G \rVert}{R} \geq \frac{\sqrt{2 L_0 A_R}}{R} \right) = 1 - \Phi(\sqrt{2 L_0}).
   \end{equation*}
   Using that for sufficiently large $R$, \eqref{prob_bound_h} holds we consequently have
   \begin{equation*}
       \limsup_{R \to \infty} \mathbb{P}(H_R) \leq 1 - \Phi(\sqrt{2 L_0}).
   \end{equation*}
       \end{proof}
\end{lemma}

\newpage

\section{Implementation Details}\label{sec:exp_details}
Our implementation is based on the CIFAR-10 code released in the TorchCFM repository accompanying \citet{tong2024improving,tong2023simulation}. We use their CIFAR-10 I-CFM configuration and retain its U-Net architecture, data preprocessing, and principal training configuration. We modify the regression objective in the $L_{2,1}$ experiments and introduce the contamination mechanisms described in Table \ref{tab:contaminations}.
CIFAR-10 images are converted to tensors, and normalized channelwise with mean and standard deviation \(0.5\), yielding images in \([-1,1]^{3\times32\times32}\). For a batch of residuals \(r_i=v_t(x_{t,i})-u_{t,i}\), with \(D=3\cdot32\cdot32\), the squared loss is implemented as

$$ \frac{1}{BD}\sum_{i=1}^{B}\lVert r_i\rVert_2^2, $$

while the \(L_{2,1}\) loss is implemented as

$$ \frac{1}{B\sqrt D}\sum_{i=1}^{B}\lVert r_i\rVert_2. $$

No smoothing is applied to the Euclidean norm. Since the dimension-dependent normalization factors are constant they do not change the population minimizers, but they can affect gradient clipping and are therefore reported as used in training. For each of the contamination experiments, the contamination is generated online: every target image occurring in a minibatch is independently corrupted with probability \(\epsilon\). Whether an image gets corrupted and associated noise are generated using a separate seeded random-number generator, so that matching randomization is used across objectives. We use a U-Net with 128 base channels,  channel multipliers \([1,2,2,2]\), attention at resolution \(16\), dropout $\num{0.1}$, four attention heads and 64 channels per head. Models are trained for 400,000 optimizer updates using Adam with learning rate \(2\cdot10^{-4}\), batch size $\num{128}$, 5,000 linear warm-up steps, global gradient-norm clipping at 1.0, and EMA decay $\num{0.9999}$.

For FID evaluation, we use the CIFAR-10 evaluation script provided in the TorchCFM repository accompanying \citet{tong2024improving,tong2023simulation}, with only the modifications required to load and evaluate our trained models. We otherwise retain the original evaluation procedure and use the same pipeline for both the squared and $L_{2,1}$ objectives.

\newpage
\section{Example Images}

\begin{figure}[ht]
    \centering

    \begin{subfigure}[t]{0.235\textwidth}
        \centering
        \includegraphics[width=\linewidth]{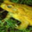}
        \caption{Clean reference}
        \label{fig:cont_random}
    \end{subfigure}\hfill
    \begin{subfigure}[t]{0.235\textwidth}
        \centering
        \includegraphics[width=\linewidth]{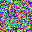}
        \caption{Random noise}
        \label{fig:cont_gaussian}
    \end{subfigure}\hfill
    \begin{subfigure}[t]{0.235\textwidth}
        \centering
        \includegraphics[width=\linewidth]{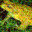}
        \caption{Gaussian noise}
        \label{fig:cont_saltpepper}
    \end{subfigure}\hfill
    \begin{subfigure}[t]{0.235\textwidth}
        \centering
        \includegraphics[width=\linewidth]{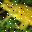}
        \caption{Salt-and-Pepper}
        \label{fig:cont_patch}
    \end{subfigure}

    \par\medskip

    \begin{subfigure}[t]{0.235\textwidth}
        \centering
        \includegraphics[width=\linewidth]{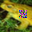}
        \caption{Patch noise}
        \label{fig:cont_pattern}
    \end{subfigure}\hfill
    \begin{subfigure}[t]{0.235\textwidth}
        \centering
        \includegraphics[width=\linewidth]{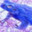}
        \caption{Inversion}
        \label{fig:cont_inversion}
    \end{subfigure}\hfill
    \begin{subfigure}[t]{0.235\textwidth}
        \centering
        \includegraphics[width=\linewidth]{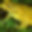}
        \caption{Gaussian blur}
        \label{fig:cont_gaussian_blur}
    \end{subfigure}\hfill
    \begin{subfigure}[t]{0.235\textwidth}
        \centering
        \includegraphics[width=\linewidth]{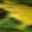}
        \caption{Motion blur}
        \label{fig:cont_motion_blur}
    \end{subfigure}

    \caption{Representative examples of the seven contamination mechanisms
    considered in our experiments. Except for random-noise replacement, all
    examples are obtained from the same clean CIFAR-10 image using the
    parameters reported in Table \ref{tab:contaminations}. }
    \label{fig:contamination_examples}
\end{figure}

\begin{figure}[ht]
    \centering

    \begin{subfigure}[t]{0.48\textwidth}
        \centering
        \includegraphics[width=\linewidth]{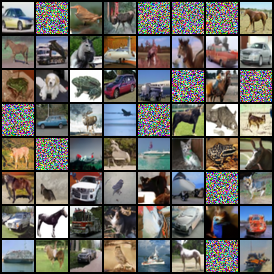}
        \caption{Squared FM}
    \end{subfigure}
    \hfill
    \begin{subfigure}[t]{0.48\textwidth}
        \centering
        \includegraphics[width=\linewidth]{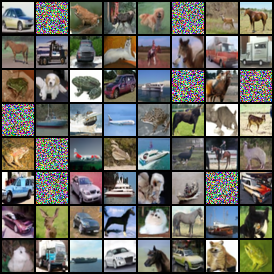}
        \caption{$L_{2,1}$ FM}
    \end{subfigure}

    \caption{
        Generated samples for Random noise with contamination level
        $\epsilon=0.15$. The same source samples are used for squared
        and $L_{2,1}$ Flow Matching.
    }
    \label{fig:CONTAMINATION_samples1}
\end{figure}

\begin{figure}[ht]
    \centering

    \begin{subfigure}[t]{0.48\textwidth}
        \centering
        \includegraphics[width=\linewidth]{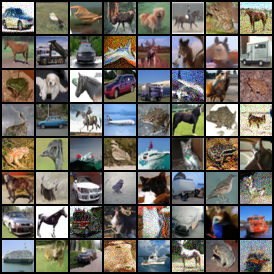}
        \caption{Squared FM}
    \end{subfigure}
    \hfill
    \begin{subfigure}[t]{0.48\textwidth}
        \centering
        \includegraphics[width=\linewidth]{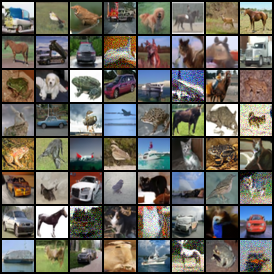}
        \caption{$L_{2,1}$ FM}
    \end{subfigure}

    \caption{
        Generated samples for Gaussian noise with contamination level
        $\epsilon=0.15$. The same source samples are used for squared
        and $L_{2,1}$ Flow Matching.
    }
    \label{fig:CONTAMINATION_samples2}
\end{figure}

\begin{figure}[ht]
    \centering

    \begin{subfigure}[t]{0.48\textwidth}
        \centering
        \includegraphics[width=\linewidth]{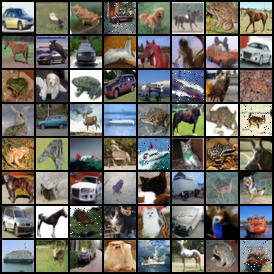}
        \caption{Squared FM}
    \end{subfigure}
    \hfill
    \begin{subfigure}[t]{0.48\textwidth}
        \centering
        \includegraphics[width=\linewidth]{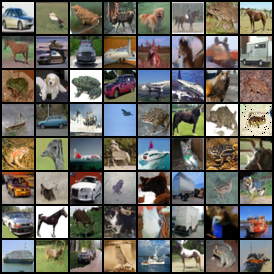}
        \caption{$L_{2,1}$ FM}
    \end{subfigure}

    \caption{
        Generated samples for Salt-and-Pepper noise with contamination level
        $\epsilon=0.15$. The same source samples are used for squared
        and $L_{2,1}$ Flow Matching.
    }
    \label{fig:CONTAMINATION_samples3}
\end{figure}

\begin{figure}[ht]
    \centering

    \begin{subfigure}[t]{0.48\textwidth}
        \centering
        \includegraphics[width=\linewidth]{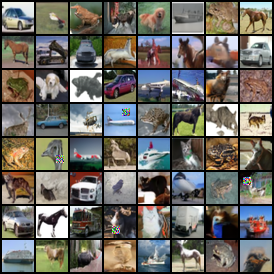}
        \caption{Squared FM}
    \end{subfigure}
    \hfill
    \begin{subfigure}[t]{0.48\textwidth}
        \centering
        \includegraphics[width=\linewidth]{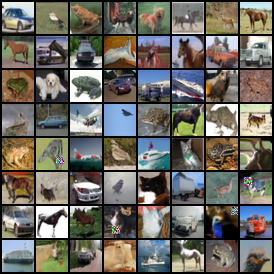}
        \caption{$L_{2,1}$ FM}
    \end{subfigure}

    \caption{
        Generated samples for Patch noise with contamination level
        $\epsilon=0.15$. The same source samples are used for squared
        and $L_{2,1}$ Flow Matching.
    }
    \label{fig:CONTAMINATION_samples4}
\end{figure}

\begin{figure}[ht]
    \centering

    \begin{subfigure}[t]{0.48\textwidth}
        \centering
        \includegraphics[width=\linewidth]{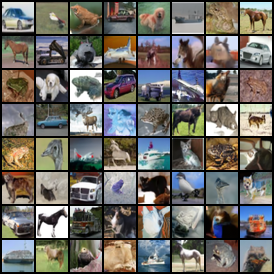}
        \caption{Squared FM}
    \end{subfigure}
    \hfill
    \begin{subfigure}[t]{0.48\textwidth}
        \centering
        \includegraphics[width=\linewidth]{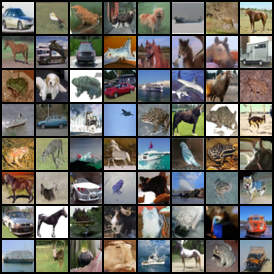}
        \caption{$L_{2,1}$ FM}
    \end{subfigure}

    \caption{
        Generated samples for Inversion with contamination level
        $\epsilon=0.15$. The same source samples are used for squared
        and $L_{2,1}$ Flow Matching.
    }
    \label{fig:CONTAMINATION_samples5}
\end{figure}

\begin{figure}[ht]
    \centering

    \begin{subfigure}[t]{0.48\textwidth}
        \centering
        \includegraphics[width=\linewidth]{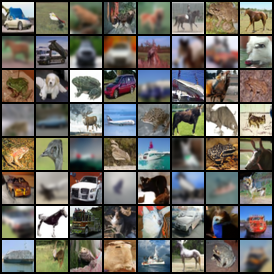}
        \caption{Squared FM}
    \end{subfigure}
    \hfill
    \begin{subfigure}[t]{0.48\textwidth}
        \centering
        \includegraphics[width=\linewidth]{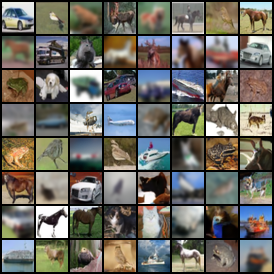}
        \caption{$L_{2,1}$ FM}
    \end{subfigure}

    \caption{
        Generated samples for Gaussian blur with contamination level
        $\epsilon=0.15$. The same source samples are used for squared
        and $L_{2,1}$ Flow Matching.
    }
    \label{fig:CONTAMINATION_samples6}
\end{figure}

\begin{figure}[ht]
    \centering

    \begin{subfigure}[t]{0.48\textwidth}
        \centering
        \includegraphics[width=\linewidth]{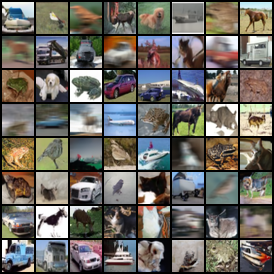}
        \caption{Squared FM}
    \end{subfigure}
    \hfill
    \begin{subfigure}[t]{0.48\textwidth}
        \centering
        \includegraphics[width=\linewidth]{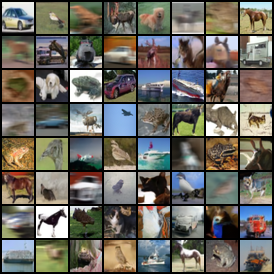}
        \caption{$L_{2,1}$ FM}
    \end{subfigure}

    \caption{
        Generated samples for Motion blur with contamination level
        $\epsilon=0.15$. The same source samples are used for squared
        and $L_{2,1}$ Flow Matching.
    }
    \label{fig:CONTAMINATION_samples7}
\end{figure}

\clearpage

\bibliography{iclr2027_conference}

@book{bookVillaniTransport,
  author    = {C{\'e}dric Villani},
  title     = {Optimal Transport: Old and New},
  series    = {Grundlehren der mathematischen Wissenschaften},
  volume    = {338},
  publisher = {Springer Berlin, Heidelberg},
  year      = {2009},
  doi       = {10.1007/978-3-540-71050-9}
}

@inproceedings{
liu2022flowstraightfastlearning,
title={Flow Straight and Fast: Learning to Generate and Transfer Data with Rectified Flow},
author={Xingchao Liu and Chengyue Gong and Qiang Liu},
booktitle={The Eleventh International Conference on Learning Representations },
year={2023},
url={https://openreview.net/forum?id=XVjTT1nw5z}
}

@inproceedings{
lipman2023flowmatchinggenerativemodeling,
title={Flow Matching for Generative Modeling},
author={Yaron Lipman and Ricky T. Q. Chen and Heli Ben-Hamu and Maximilian Nickel and Matthew Le},
booktitle={The Eleventh International Conference on Learning Representations },
year={2023},
url={https://openreview.net/forum?id=PqvMRDCJT9t}
}

@misc{lipman2024flowmatchingguidecode,
      title={Flow Matching Guide and Code}, 
      author={Yaron Lipman and Marton Havasi and Peter Holderrieth and Neta Shaul and Matt Le and Brian Karrer and Ricky T. Q. Chen and David Lopez-Paz and Heli Ben-Hamu and Itai Gat},
      year={2024},
      eprint={2412.06264},
      archivePrefix={arXiv},
      primaryClass={cs.LG},
      url={https://arxiv.org/abs/2412.06264}, 
}

@article{albergo2025stochasticinterpolantsunifyingframework,
  title={Stochastic interpolants: A unifying framework for flows and diffusions},
  author={Albergo, Michael and Boffi, Nicholas M and Vanden-Eijnden, Eric},
  journal={Journal of Machine Learning Research},
  volume={26},
  number={209},
  pages={1--80},
  year={2025}
}

@inproceedings{peyré2025optimaldiffusiontransportsmachine,
author = {Gabriel Peyré},
title = {Optimal and Diffusion Transports in Machine Learning},
booktitle = {Proceedings of the International Congress of Mathematicians 2026 - Volume 7: Invited Lectures (Sections 15–20)},
pages = {110--128},
doi = {10.1137/25M1803966},
URL = {https://epubs.siam.org/doi/abs/10.1137/25M1803966},
year={2026},
publisher={Society for Industrial and Applied Mathematics}
}

@InProceedings{pooladian2023multisampleflowmatchingstraightening,
  title = 	 {Multisample Flow Matching: Straightening Flows with Minibatch Couplings},
  author =       {Pooladian, Aram-Alexandre and Ben-Hamu, Heli and Domingo-Enrich, Carles and Amos, Brandon and Lipman, Yaron and Chen, Ricky T. Q.},
  booktitle = 	 {Proceedings of the 40th International Conference on Machine Learning},
  pages = 	 {28100--28127},
  year = 	 {2023},
  editor = 	 {Krause, Andreas and Brunskill, Emma and Cho, Kyunghyun and Engelhardt, Barbara and Sabato, Sivan and Scarlett, Jonathan},
  volume = 	 {202},
  series = 	 {Proceedings of Machine Learning Research},
  month = 	 {23--29 Jul},
  publisher =    {PMLR},
  url = 	 {https://proceedings.mlr.press/v202/pooladian23a.html}
}

@article{10.1214/aos/1176350511,
author = {P. Milasevic and G. R. Ducharme},
title = {{Uniqueness of the Spatial Median}},
volume = {15},
journal = {The Annals of Statistics},
number = {3},
publisher = {Institute of Mathematical Statistics},
pages = {1332--1333},
year = {1987},
doi = {10.1214/aos/1176350511},
URL = {https://doi.org/10.1214/aos/1176350511}
}

@InProceedings{tong2023simulation,
  title = 	 {Simulation-Free {S}chrödinger Bridges via Score and Flow Matching},
  author =       {Tong, Alexander Y. and Malkin, Nikolay and Fatras, Kilian and Atanackovic, Lazar and Zhang, Yanlei and Huguet, Guillaume and Wolf, Guy and Bengio, Yoshua},
  booktitle = 	 {Proceedings of The 27th International Conference on Artificial Intelligence and Statistics},
  pages = 	 {1279--1287},
  year = 	 {2024},
  editor = 	 {Dasgupta, Sanjoy and Mandt, Stephan and Li, Yingzhen},
  volume = 	 {238},
  series = 	 {Proceedings of Machine Learning Research},
  month = 	 {02--04 May},
  publisher =    {PMLR},
  url = 	 {https://proceedings.mlr.press/v238/tong24a.html}
}

@misc{khrapov2024improvingdiffusionmodelssdatacorruption,
      title={Improving Diffusion Models's Data-Corruption Resistance using Scheduled {Pseudo-Huber} Loss}, 
      author={Artem Khrapov and Vadim Popov and Tasnima Sadekova and Assel Yermekova and Mikhail Kudinov},
      year={2024},
      eprint={2403.16728},
      archivePrefix={arXiv},
      primaryClass={cs.AI},
      url={https://arxiv.org/abs/2403.16728}, 
}

@article{Minsker_2015,
   title={Geometric median and robust estimation in {Banach} spaces},
   volume={21},
   ISSN={1350-7265},
   url={http://dx.doi.org/10.3150/14-BEJ645},
   pages  = {2308--2335},
   DOI={10.3150/14-bej645},
   number={4},
   journal={Bernoulli},
   publisher={Bernoulli Society for Mathematical Statistics and Probability},
   author={Minsker, Stanislav},
   year={2015} }

@article{10.1093/biomet/35.3-4.414,
    author = {Haldane, J. B. S.},
    title = {Note on the median of a multivariate distribution},
    journal = {Biometrika},
    volume = {35},
    number = {3-4},
    pages = {414--417},
    year = {1948},
    issn = {0006-3444},
    doi = {10.1093/biomet/35.3-4.414},
    url = {https://doi.org/10.1093/biomet/35.3-4.414},
    eprint = {https://academic.oup.com/biomet/article-pdf/35/3-4/414/785927/35-3-4-414.pdf},
}

@article{1f6117c1-50e3-3658-911b-2cdfe87c4f25,
 ISSN = {03067734, 17515823},
 URL = {http://www.jstor.org/stable/1403809},
 author = {Christopher G. Small},
 journal = {International Statistical Review / Revue Internationale de Statistique},
 number = {3},
 pages = {263--277},
 publisher = {[Wiley, International Statistical Institute (ISI)]},
 title = {A Survey of Multidimensional Medians},
 urldate = {2026-08-04},
 volume = {58},
 year = {1990}
}

@article{doi:10.1137/23M1592420,
author = {Minsker, Stanislav and Strawn, Nate},
title = {The Geometric Median and Applications to Robust Mean Estimation},
journal = {SIAM Journal on Mathematics of Data Science},
volume = {6},
number = {2},
pages = {504--533},
year = {2024},
doi = {10.1137/23M1592420},

URL = { 
    
        https://doi.org/10.1137/23M1592420
    
    

},
eprint = { 
    
        https://doi.org/10.1137/23M1592420
    
    

}
}

@article{article,
author = {Lopuhaä, Hendrik and Rousseeuw, Peter},
year = {1991},
pages = {229--248},
title = {Breakdown Points of Affine Equivariant Estimators of Multivariate Location and Covariance Matrices},
volume = {19},
journal = {The Annals of Statistics},
doi = {10.1214/aos/1176347978}
}

@InProceedings{pmlr-v258-schwank25a,
  title = 	 {Robust Score Matching},
  author =       {Schwank, Richard and McCormack, Andrew and Drton, Mathias},
  booktitle = 	 {Proceedings of The 28th International Conference on Artificial Intelligence and Statistics},
  pages = 	 {1234--1242},
  year = 	 {2025},
  editor = 	 {Li, Yingzhen and Mandt, Stephan and Agrawal, Shipra and Khan, Emtiyaz},
  volume = 	 {258},
  series = 	 {Proceedings of Machine Learning Research},
  month = 	 {03--05 May},
  publisher =    {PMLR},
  url = 	 {https://proceedings.mlr.press/v258/schwank25a.html}
}

@misc{l1sampleflow,
      title={L1 Sample Flow for Efficient Visuomotor Learning}, 
      author={Weixi Song and Zhetao Chen and Tao Xu and Xianchao Zeng and Xinyu Zhou and Lixin Yang and Donglin Wang and Cewu Lu and Yong-Lu Li},
      year={2025},
      eprint={2511.17898},
      archivePrefix={arXiv},
      primaryClass={cs.RO},
      url={https://arxiv.org/abs/2511.17898}, 
}

@misc{li2026robustdiffusionmodelsdivergenceinduced,
      title={Robust Diffusion Models via Divergence-Induced Weighted Denoising}, 
      author={Lei Li and Yuexiao Dong},
      year={2026},
      eprint={2606.22521},
      archivePrefix={arXiv},
      primaryClass={stat.ML},
      url={https://arxiv.org/abs/2606.22521}, 
}

@techreport{krizhevsky2009learningML,
  author      = {Alex Krizhevsky},
  title       = {Learning Multiple Layers of Features from Tiny Images},
  institution = {University of Toronto},
  year        = {2009},
  url         = {https://www.cs.toronto.edu/~kriz/learning-features-2009-TR.pdf}
}

@article{tong2024improving,
title={Improving and generalizing flow-based generative models with minibatch optimal transport},
author={Alexander Tong and Kilian Fatras and Nikolay Malkin and Guillaume Huguet and Yanlei Zhang and Jarrid Rector-Brooks and Guy Wolf and Yoshua Bengio},
journal={Transactions on Machine Learning Research},
issn={2835-8856},
year={2024},
url={https://openreview.net/forum?id=CD9Snc73AW},
}

@inproceedings{
albergo2023buildingnormalizingflowsstochastic,
title={Building Normalizing Flows with Stochastic Interpolants},
author={Michael Samuel Albergo and Eric Vanden-Eijnden},
booktitle={The Eleventh International Conference on Learning Representations },
year={2023},
url={https://openreview.net/forum?id=li7qeBbCR1t}
}

@inproceedings{NEURIPS2018_69386f6b,
 author = {Chen, Ricky T. Q. and Rubanova, Yulia and Bettencourt, Jesse and Duvenaud, David K},
 booktitle = {Advances in Neural Information Processing Systems},
 editor = {S. Bengio and H. Wallach and H. Larochelle and K. Grauman and N. Cesa-Bianchi and R. Garnett},
 pages = {},
 publisher = {Curran Associates, Inc.},
 title = {Neural Ordinary Differential Equations},
 url = {https://proceedings.neurips.cc/paper_files/paper/2018/file/69386f6bb1dfed68692a24c8686939b9-Paper.pdf},
 volume = {31},
 year = {2018}
}

@inproceedings{
grathwohl2018ffjord,
title={Scalable Reversible Generative Models with Free-form Continuous Dynamics},
author={Will Grathwohl and Ricky T. Q. Chen and Jesse Bettencourt and David Duvenaud},
booktitle={International Conference on Learning Representations},
year={2019},
url={https://openreview.net/forum?id=rJxgknCcK7},
}

@book{rob_outlier,
    author = {Rousseeuw, Peter and Leroy, Annick},
    year = {1987},
    pages = {358--364},
    title = {Robust Regression and Outlier Detection},
    publisher = {John Wiley \& Sons},
    doi = {10.1002/0471725382}
}

@article{DORMAND198019,
title = {A family of embedded {Runge-Kutta} formulae},
journal = {Journal of Computational and Applied Mathematics},
volume = {6},
number = {1},
pages = {19--26},
year = {1980},
issn = {0377-0427},
doi = {10.1016/0771-050X(80)90013-3},
url = {https://www.sciencedirect.com/science/article/pii/0771050X80900133},
author = {J.R. Dormand and P.J. Prince}
}

@inproceedings{NIPS2017_8a1d6947,
 author = {Heusel, Martin and Ramsauer, Hubert and Unterthiner, Thomas and Nessler, Bernhard and Hochreiter, Sepp},
 booktitle = {Advances in Neural Information Processing Systems},
 editor = {I. Guyon and U. Von Luxburg and S. Bengio and H. Wallach and R. Fergus and S. Vishwanathan and R. Garnett},
 pages = {},
 publisher = {Curran Associates, Inc.},
 title = {{GAN}s Trained by a Two Time-Scale Update Rule Converge to a Local {Nash} Equilibrium},
 url = {https://proceedings.neurips.cc/paper_files/paper/2017/file/8a1d694707eb0fefe65871369074926d-Paper.pdf},
 volume = {30},
 year = {2017}
}

@article{10.1214/aoms/1177703732,
author = {Peter J. Huber},
title = {{Robust Estimation of a Location Parameter}},
volume = {35},
journal = {The Annals of Mathematical Statistics},
number = {1},
publisher = {Institute of Mathematical Statistics},
pages = {73 -- 101},
year = {1964},
doi = {10.1214/aoms/1177703732},
URL = {https://doi.org/10.1214/aoms/1177703732}
}

\end{appendices}
\end{document}